\documentclass[journal]{IEEEtran}

\usepackage{graphicx}
\usepackage{amsmath,amsfonts}
\usepackage{float}  
\usepackage{subfigure} 
\usepackage{amssymb}
\usepackage{booktabs}
\usepackage{multirow}
\usepackage{diagbox}
\usepackage{url}
\usepackage{color}
\usepackage{algorithmic}
\usepackage{algorithm}
\usepackage{array}
\usepackage{textcomp}
\usepackage{stfloats}
\usepackage{verbatim}
\usepackage{graphicx}
\usepackage{cite}
\usepackage{flushend}
\def\BibTeX{{\rm B\kern-.05em{\sc i\kern-.025em b}\kern-.08em
    T\kern-.1667em\lower.7ex\hbox{E}\kern-.125emX}}
\usepackage{balance}

\begin{document}
\title{BACTrack: Building Appearance Collection \\for Aerial Tracking}
\author{Xincong Liu, Tingfa Xu$^{\ast}$, Ying Wang, Zhinong Yu, Xiaoying Yuan, Haolin Qin, and Jianan Li$^{\ast}$\thanks{*Corresponding author}.
\IEEEcompsocitemizethanks{\IEEEcompsocthanksitem 
Xincong Liu, Tingfa Xu, Ying Wang, Zhinong Yu, Xiaoying Yuan, Haolin Qin, and Jianan Li are with Beijing Institute of Technology, Beijing 100081, China (e-mail: \{3120210514, ciom\_xtf1, 3120215325, znyu, 3120210585, 3120225333, lijianan\}@bit.edu.cn).
\IEEEcompsocthanksitem 
Jianan Li and Tingfa Xu are also with the Key Laboratory of Photoelectronic Imaging Technology and System, Ministry of Education of China, Beijing 100081, China 
\IEEEcompsocthanksitem 
Tingfa Xu is also with  Chongqing Innovation Center, Beijing Institute of Technology, Chongqing 401135, China.}}

\maketitle

\begin{abstract}
Siamese network-based trackers have shown remarkable success in aerial tracking. Most previous works, however, usually perform template matching only between the initial template and the search region and thus fail to deal with rapidly changing targets that often appear in aerial tracking. As a remedy, this work presents Building Appearance Collection Tracking (BACTrack). This simple yet effective tracking framework builds a dynamic collection of target templates online and performs efficient multi-template matching to achieve robust tracking. Specifically, BACTrack mainly comprises a Mixed-Temporal Transformer (MTT) and an appearance discriminator. The former is responsible for efficiently building relationships between the search region and multiple target templates in parallel through a mixed-temporal attention mechanism. At the same time, the appearance discriminator employs an online adaptive template-update strategy to ensure that the collected multiple templates remain reliable and diverse, allowing them to closely follow rapid changes in the target's appearance and suppress background interference during tracking.  
Extensive experiments show that our BACTrack achieves top performance on four challenging aerial tracking benchmarks while maintaining an impressive speed of over 87 FPS on a single GPU.
Speed tests on embedded platforms also validate our potential suitability for deployment on UAV platforms.

\end{abstract}


\section{Introduction}
\IEEEPARstart{V}{isual} object tracking is the task of tracking targets throughout a video sequence, given only the target annotation in the initial frame.
As an essential branch of object tracking, unmanned aerial vehicles (UAV) tracking has received increasing attention for its applications in aerial cinematography, geographical survey, \emph{etc}\cite{bonatti2019towards,app1}.
UAV tracking, pivotal for applications such as security enforcement, presents distinct challenges compared to conventional object tracking\cite{han2022comprehensive,MDOT}:
(1) The expansive aerial field of vision introduces numerous distractors and background clutter, complicating the tracking process. 
(2) Variations in the UAV's perspective and the target's rapid movement result in substantial alterations in the target's visual characteristics.
(3) The aerial platform's limited computational capabilities, often lacking GPU acceleration, poses a substantial challenge in developing tracking algorithms that balance low complexity with high accuracy \cite{HiFT}.  
Therefore, despite the breakthroughs in UAV tracking algorithms, accurate and efficient UAV tracking remains challenging.
\begin{figure}
\centering
\subfigure[Overview of BACTrack]{
\label{fig:Motivation1}
\includegraphics[width=0.98\linewidth]{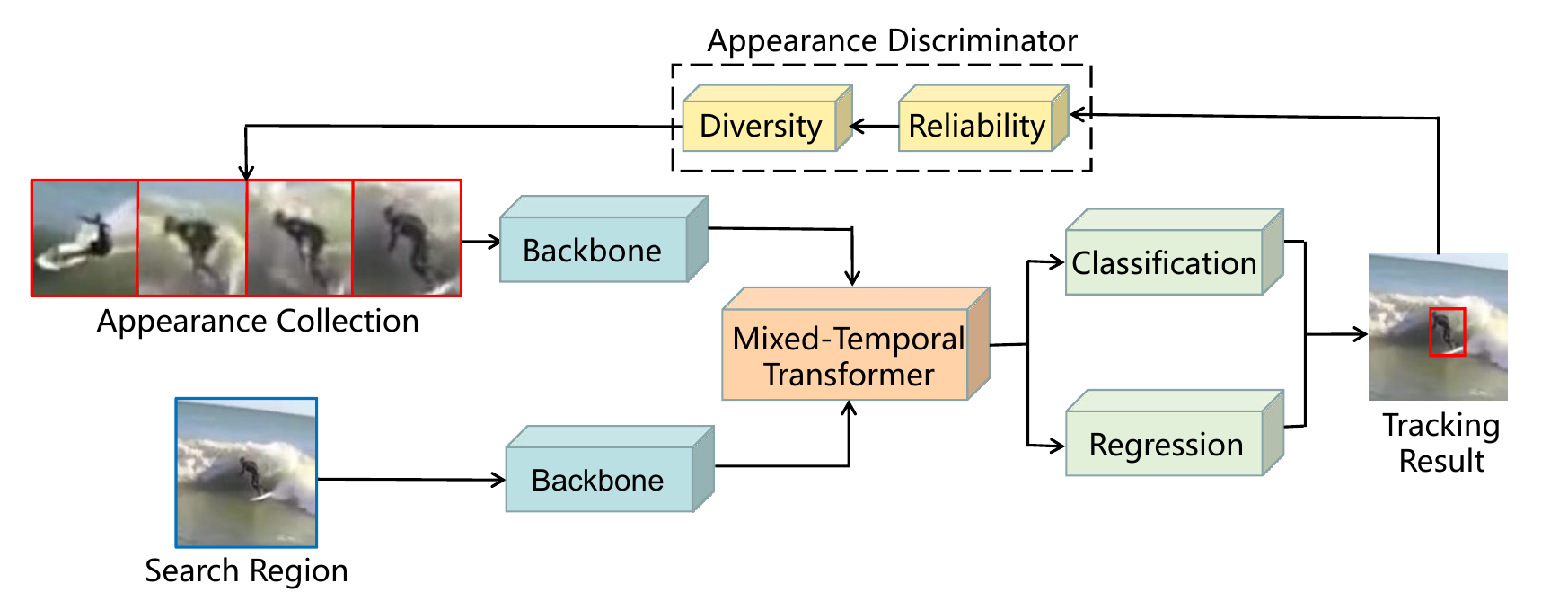}}
\subfigure[Accuracy-speed trade-off on DTB70]{
\label{fig:Motivation2}
\hspace{12mm}\includegraphics[width=0.75\linewidth]{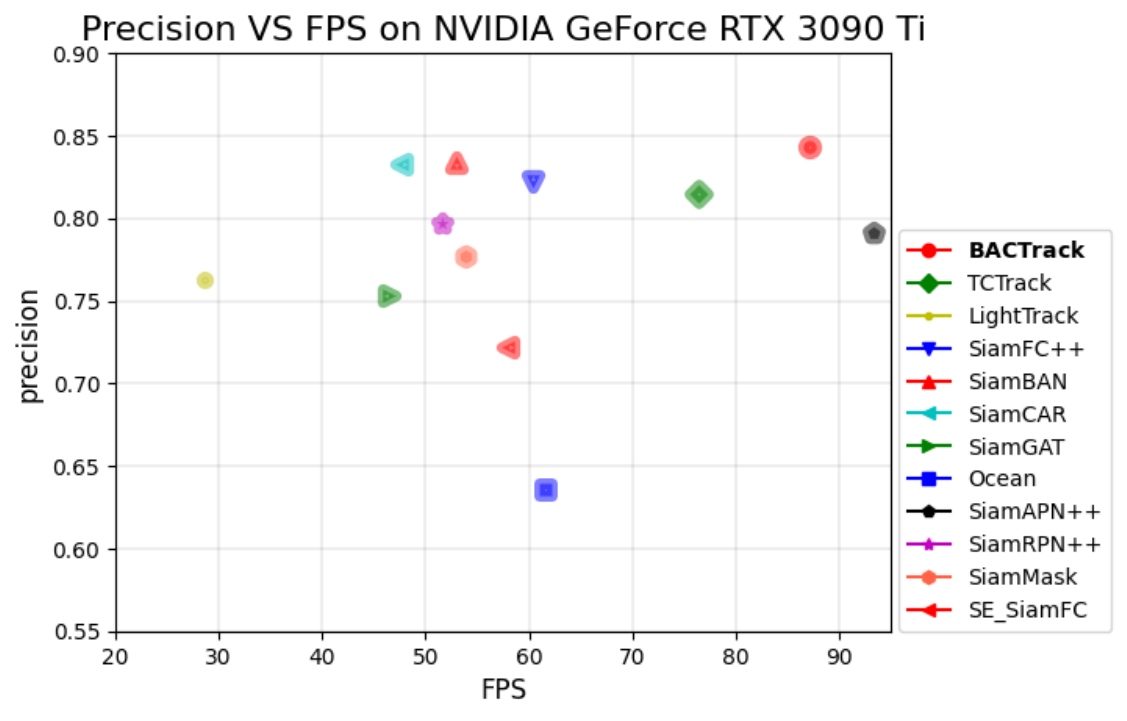}}
\caption{\textbf{(a) The overall framework of BACTrack. (b) Accuracy-speed trade-off on DTB70.}
The speed evaluation platform is NVIDIA GeForce RTX 3090 Ti.
Thanks to the appearance collection, our BACTrack achieves competitive tracking performance with impressive tracking speed.}
\label{fig:Motivation}
\end{figure}

As a one-shot problem, the initial frame provides accurate location but limited appearance information of the target, making it difficult to cope with the subsequent complex changes in appearance and background disturbances. 
Therefore, many aerial tracking algorithms have integrated temporal contexts into the frameworks mechanisms of \emph{temporal-dispersion}\cite{DSiam,TMT,Stmtrack} and \emph{temporal-accumulate}\cite{TCTrack,updatenet}, leading to notable improvement in performance.
Algorithms based on \emph{temporal-dispersion} exploit discrete temporal information in tracking sequences, either by replacing the template or integrating new features to improve characterization capabilities.
However, temporal template may inadequately represent target's appearance, leading to tracker drift during long-time tracking and rapid target deformation.
In turn, algorithms based on \emph{temporal-accumulate} accumulate historical frames during the tracking process.
Benefiting from the rich historical contexts, these algorithms perform better but inevitably require more computational resources, thus impeding further applications.

Accordingly, in order to address the challenges in aerial tracking outlined earlier, we aim to present a lightweight aerial tracking framework named BACTrack to build a dynamic collection of online target templates.
As shown in Fig.~\ref{fig:Motivation1}, BACTrack consists of a feature extraction backbone, a Mixed-Temporal Transformer (MTT) module, a prediction head, and an appearance discriminator (AD).
The weight-shared backbone extracts multiple template features and the current search feature, which are simultaneously fused by the MTT.
The prediction head then processes the mixed-temporal features for target localization. Subsequently, the  output from each frame is fed to the appearance discriminator, which assesses the necessity for updates to the appearance collection.

The templates in the appearance collection are required to be both reliable and diverse. Accordingly, the online appearance discriminator adopts an adaptive template-update strategy.
The strategy consists of two parts: 
For \textbf{reliability}, we employ an adaptive confidence score threshold to ensure that the updated template accurately represent the target rather than the background. This effectively mitigates issue of background interference in aerial tracking.
For \textbf{diversity}, we employ the Structural Similarity Index Measure (SSIM)~\cite{SSIM} to reflect the similarity between templates.
During the tracking process, we retain adequate temporal information within the appearance collection so that it can adapt to the current state of the target and effectively address the challenge of deformation from an airborne view.
Hence, the refined appearance collection achieves a well-balanced complement to the initial template.

Based on the appearance collection, we propose a novel approach to temporal feature fusion.
Previous research on the grouping of attention heads has shown to be effective. Specifically, SSA\cite{ren2022shunted} divides attention heads into multiple groups to aggregate image features of different granularity. MsSVT\cite{dong2022mssvt} extends window-based attention on 3D voxels. Therefore, we explore the application of this idea to temporal contexts and propose a mixed-temporal attention (MTA) mechanism in the MTT. 
MTA allocates the templates from the appearance collection to different heads, facilitating efficient multi-template matching. Consequently, the MTT focuses on the relationships between the search region and multiple target template in parallel by grouping the heads in an attention layer. 
As a result, BACTrack performs efficient multi-template matching to fully perceive rapid target deformation and achieve robust long-time tracking while incurring little computational consumption.

We conducted an extensive evaluation of our BACTrack on several aerial tracking benchmarks, including DTB70\cite{DTB70}, UAV123\cite{uav123}, UAV123@10fps\cite{uav123}, and UAVTrack112\_L\cite{fu2021onboard112}. 
It demonstrated superior performance over the top-performing TCTrack\cite{TCTrack} by $1\%$ in area-under-the-curve (AUC) and $3.4\%$ in precision on DTB70, running at an impressive real-time speed of $87.2$ frames per second (FPS) on PC (Fig.~\ref{fig:Motivation2}), achieving an ideal trade-off between accuracy and speed.
Furthermore, when deployed on the NVIDIA Jetson AGX Xavier, a UAV test platform, our BACTrack maintains a remarkable speed of 28 FPS, highlighting its suitability for deployment on UAV platforms.

The main contributions of this work are summarized as follows:
\begin{itemize}
\item We propose a Mixed-Temporal Transformer (MTT) module, which is designed to enhance feature fusion by consolidating multi-frame feature integration into a single attention layer. This enhancement is achieved by introducing a novel mixed-temporal attention mechanism, which is specifically designed to augment the accuracy in estimating the current state of the target.
\item We present an appearance discriminator (AD) equipped with an online adaptive template-update strategy to ensure that collected templates closely follow the target's appearance degradation and suppress the background during the tracking process.
\item We propose BACTrack, an innovative aerial tracking framework that introduces temporal information by constructing a historical collection of target appearances. This collection's incorporation significantly enhances the tracker's robustness against variations in the target's appearance.
\end{itemize}

\section{Related Work}
\label{sec:Related Work}

\subsection{Aerial Tracking}
Aerial tracking has recently gained significant attention due to the flexibility and convenience offered by UAVs.
In this domain, two prevalent methods are particularly notable: Discriminative Correlation Filters (DCF) and deep Siamese Networks.
DCF-based trackers\cite{DCF1,DCF2ATOM,AutoTrack} are widely adopted in UAV tracking due to their notable attributes, including low computational complexity and real-time processing capabilities\cite{ARCF,han2019state,deng2021learning}. However, despite their advantages, DCF-based trackers often face difficulties in complex scenes, such as occlusion.
Conversely, trackers based on Siamese networks have achieved significant performance by utilizing deep features\cite{SiamFC,dong2022adaptive}. These networks are designed to learn the similarity between the template and search region through end-to-end offline training.
For example, the Siamese Residual Network\cite{fan2023siamese} proposes an identity branch and a residual branch to handle complex scenarios and dramatic target appearance variations.
Several studies have focused on optimizing efficiency, achieving real-time tracking on mobile platforms\cite{HiFT,siamapn++}. 
Additionally, significant efforts have been made to address hyperparameter optimization\cite{dong2019dynamical} and occlusion problems\cite{dong2016occlusion}.
In this work, we explore Siamese tracking for efficient aerial tracking, building upon the models mentioned above and balancing accuracy and speed. 

\subsection{Vision transformers}
Transformer was originally developed for natural language processing tasks\cite{attention} and has achieved notable success in the field of computer vision\cite{vt2,liu2021swinvt3,transT}. 
Swin-transformer\cite{liu2021swinvt3} introduces a hierarchical sliding window structure, which confines self-focusing to local windows and allows cross-window connections for efficiency.
SSA\cite{ren2022shunted} divides attention heads into multiple groups to aggregate image features of different granularity, capable of capturing multi-scale objects.
MsSVT\cite{dong2022mssvt} extends window-based attention on 3D voxels by introducing scale-aware attention learning bolstered by novel sampling strategies applied to the queries and the keys.
TransT\cite{transT}, TMT\cite{TMT}, and others\cite{Stark,trackformer,mixformer,TransMDOT,SFTransT} propose various transformer structures applied better to object tracking tasks. 
STARK\cite{Stark} leverages the transformer to integrate spatial and temporal information.
SFTransT\cite{SFTransT} fuses cross-scale representations from various transformer layers to preserve multi-stage information from extracted features.
Our study extends attention grouping in tracking, drawing inspiration from the aforementioned models. We introduce multi-frame templates in the feature fusion of Siamese trackers, equipped with a novel template-update strategy to improve the robustness and accuracy.

\subsection{Feature Fusion in Tracking}
Most Siamese trackers view object tracking as a template-matching process between the template and the search region. Consequently, various feature fusion strategies have been explored.
To effectively capture the correlation between the template and the search region features, most methods employ the variants of cross-correlation, including naive cross-correlation\cite{SiamFC}, depth-wise cross-correlation\cite{siamrpn++,ocean}, and point-wise cross-correlation\cite{alpha}. MMNet\cite{liu2022learning} designs a fine-grained aware network to get the fine-grained correlation features.
Despite their commendable performance, these methods tend to capture only the local correlation while neglecting the broader global information context.
In contrast, the attention mechanism\cite{attention} has proven suitable for feature fusion tasks due to its ability to capture global relationships\cite{transT,mixformer,TaTrack}. TransT\cite{transT}, for example, introduces an attention-based feature fusion network that combines template and search region features solely using attention.
However, given that the target's appearance may undergo alterations during tracking, relying solely on the correlation of the initial template often leads to tracking drift.
To address this issue, we explore the method of simultaneous fusion of multiple templates using attention mechanisms.

\subsection{Temporal Information in Visual Tracking}
Due to the complex tracking environment, the target may undergo severe appearance changes, including significant motion and occlusion. In order to perceive the motion information of target objects, many studies have introduced temporal information into the visual tracking task, which has been proven to have a significant contribution to the robustness of the trackers\cite{Stark,TMT,Stmtrack,TCTrack,updatenet,TaTrack,yuan2023active,deng2021learning}.
STCAT\cite{han2019spatial} incorporates both the spatial and temporal constrain into the DCF framework.
In Siamese trackers, most existing methods introduce temporal information through temporal-dispersion and temporal-accumulate. 
The temporal-dispersion-based algorithms\cite{DSiam,gradnet,Stark,TMT,Stmtrack} replace the template or incorporate new features to exploit the discrete temporal information. 
DSiam\cite{DSiam} proposes a dynamic siamese network to enable adequate template updating and cluttered background suppression.
STMTrack\cite{Stmtrack} divides the historical tracking frames equally and takes the intermediate frames as temporal information.
On the contrary, the temporal-accumulate-based trackers perceive temporal information by accumulating context for each successive frame, such as UpdateNet\cite{updatenet} and TCTrack\cite{TCTrack}. For example, UpdateNet generates an optimal template by weighting the previously accumulated templates and the template of the current frame.
Due to the rich historical information, these trackers perform better but inevitably have a greater demand on computational resources. 
Our BACTrack builds an appearance collection and introduces discrete temporal information into the framework through multi-frame feature fusion, combining the advantages of both approaches.

\subsection{Multiple Object Tracking in Aerial Tracking}
Tracking multiple objects of similar appearance in aerial tracking is increasingly popular, aiming to accurately locate the trajectory of each target while maintaining their identities based on information accumulated up to the current frame. In the field of aerial tracking, multiple object tracking (MOT) presents unique challenges due to abrupt appearance changes and similar objects\cite{mot}.
There are two main paradigms to in MOT: tracking-by-detection (TBD) and joint detection embedding (JDE). The TBD\cite{zhou2023f} method associates the target detection results after they are obtained by the detector. The JDE algorithm\cite{wang2020towards} combines the detection module and the tracking module to achieve faster inference. For example, UMA\cite{yin2020unified} unifies object motion and affinity model into a single network, effectively improving the computation efficiency. 
In response to the many challenges of aerial scenes, scholars have proposed improvements in several aspects, involving optimizing network models\cite{rs13091670,STDFormer}, introducing attention mechanisms\cite{10082104,sun2020transtrack}, fusing multi-scale features\cite{9376456}, and developing lightweight network\cite{rs13040573}.
Given that aerial tracking often involves multiple objects with similar appearances, we have also validated the effectiveness of our method in scenarios with multiple targets, ensuring it can accurately distinguish and track each individual object.

\begin{figure*}
\centering
\subfigure[Overall network architecture of BACTrack]{
\label{fig:framework}
\includegraphics[width=11cm,height = 5cm]{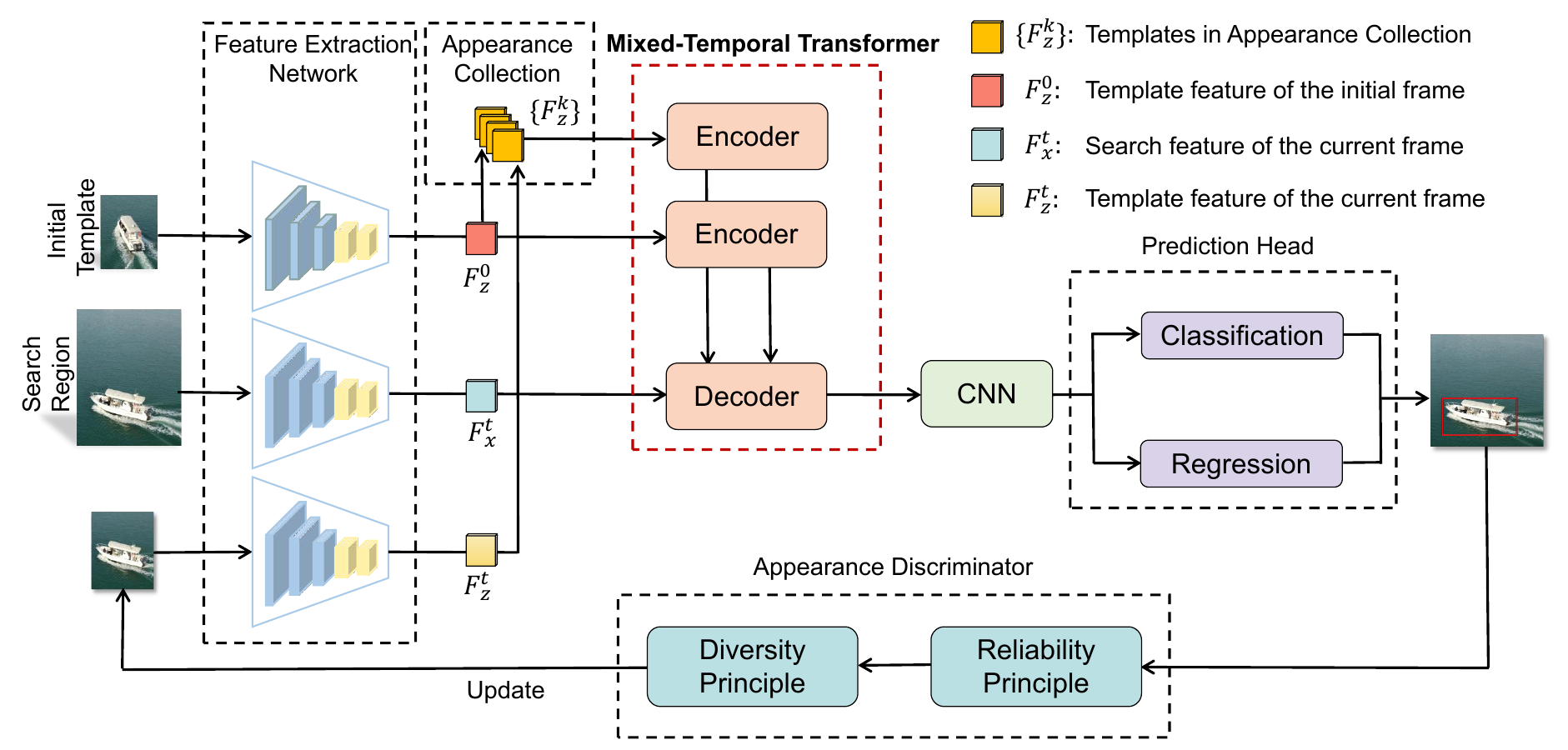}}
\subfigure[Mixed-Temporal Transformer]{
\label{fig:MTT}
\includegraphics[width=6cm,height = 5cm]{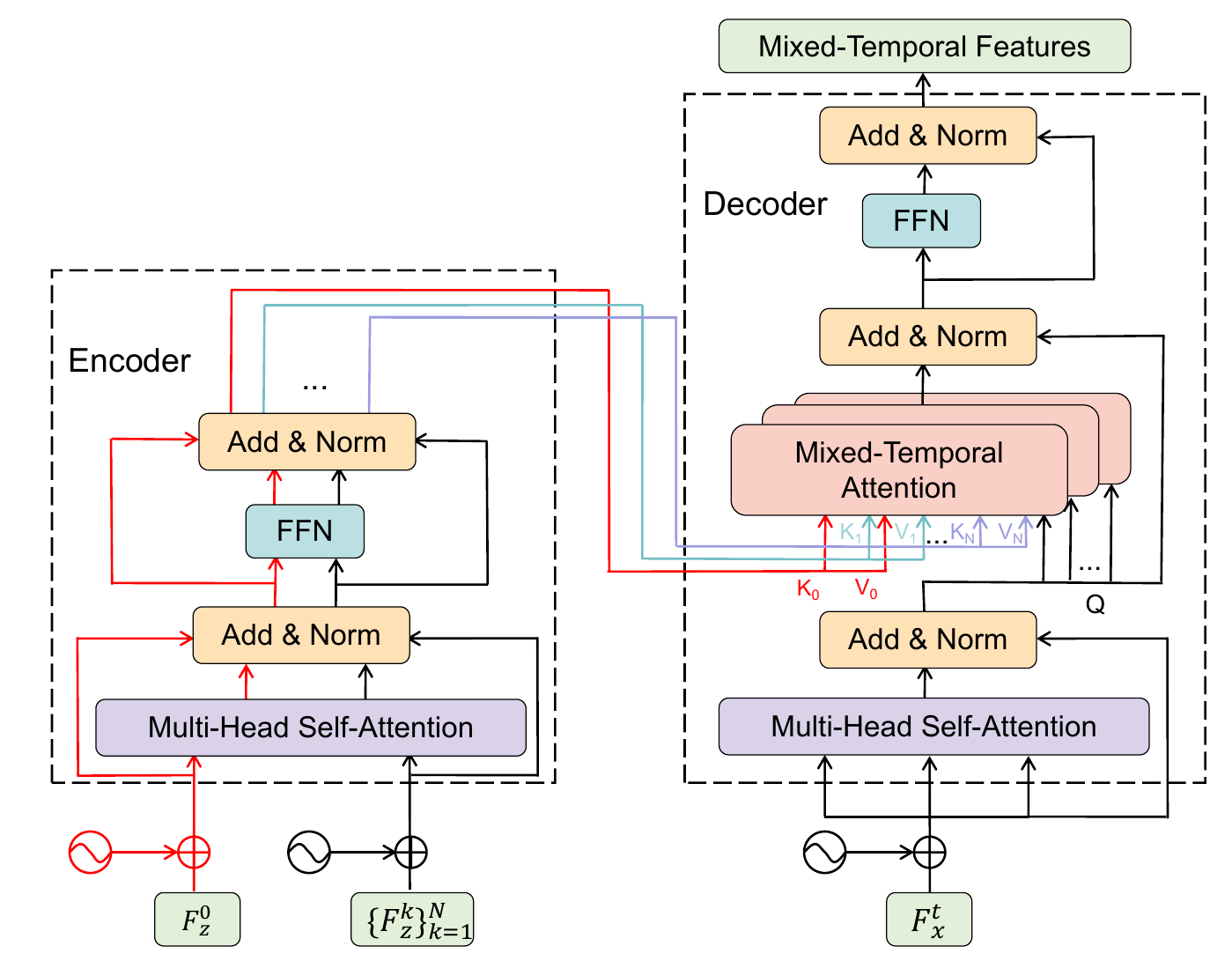}}
\caption{\textbf{(a) Overall network architecture of BACTrack.}
BACTrack consists of four components: a feature extraction network, a Mixed-Temporal Transformer (MTT) module for feature fusion, a prediction head, and an appearance discriminator (AD). 
The extracted search features of the current frame are fused with multiple target templates in parallel in MTT. 
The prediction head then processes the generated mixed-temporal features. The final tracking results are further fed to AD for template updating.
\textbf{(b) The architecture of Mixed-Temporal Transformer.} The MTT module is built with an encoder-decoder architecture, and mixed-temporal attention is proposed in the decoder for multi-template feature fusion.}
\end{figure*}

\section{Method}
The overall network architecture of BACTrack, as illustrated in Fig.~\ref{fig:framework}, is sequentially composed of four parts:
a feature extraction network, a Mixed-Temporal Transformer (MTT), a prediction head, and an Appearance Discriminator (AD). 
Initially, BACTrack extracts features through the backbone. These extracted features are then fed into the MTT, designed for effective feature fusion. The search feature is integrated with multiple template features from the appearance collection to adapt to changes in the targets' appearance.
The prediction head then processes the fused features for target localization. The final tracking results are further fed to AD, which is responsible for building and updating the appearance collection. 
Specifically, we introduce an appearance collection in the framework to memorize the templates and equip it with an appearance discriminator for updating. 
During online tracking, this collection is optimized to encompass the most reliable and representative templates.
Detailed descriptions are presented below.
\subsection{Siamese Feature Extraction}
Our baseline model is TCTrack\cite{TCTrack}, following its proposed feature extraction network, the online Temporally Adaptive Convolutional Neural Network (TAdaCNN), to generate corresponding features. This network integrates temporal contexts in the stage of feature extraction by TAdaConv\cite{tada}.
The input to the feature extraction network comprises a set of image patches: a search region image patch ${\rm x}_t\in\mathbb{R}^{\rm {H_{s}\times W_{s} \times 3}}$ and template image patches from the appearance collection.
This collection includes the initial template patch ${\rm z_0\in\mathbb{R}^{ H_{t}\times W_{t} \times 3}}$ and temporal template patches ${\rm z}_k\in\mathbb{R}^{\rm H_{t}\times W_{t} \times 3}$, where $k$ ranges from 1 to $\rm N$, and ${\rm N}$ represents the number of temporal templates.
Similar to most Siamese trackers, the process of extracting features can be formally denoted as:
\begin{equation}
\mathbf{F}_{\rm z}^0, \mathbf{F}_{\rm z}^k, \mathbf{F}_{\rm x}^t = {\rm \mathbf{\Phi}}_{\rm tada}({\rm z_0}, {\rm z}_k, {\rm x}_t),
\end{equation}
where ${\rm \mathbf{\Phi}}_{\rm tada}$ represents the online TAdaCNN.
For the $t$-th frame, the final output of the feature extraction network is represented as $\mathbf{F}_{\rm z}^t=\left\{\mathbf{F}_{\rm z}^0,\mathbf{F}_{\rm z}^1, \dots, \mathbf{F}_{\rm z}^{\rm N}\right\}$ and $\mathbf{F}_{\rm x}^t$, where $\mathbf{F}_{\rm x}^t\in\mathbb{R}^{\rm {H_{x}\times W_{x} \times C}}$ represents the search region feature of the $t$-th frame, $\mathbf{F}_{\rm z}^0\in\mathbb{R}^{\rm {H_{z}\times W_{z} \times C}}$ represents the template feature of the initial frame, and $\mathbf{F}_{\rm z}^{1}, \dots, \mathbf{F}_{\rm z}^{{\rm N}}\in\mathbb{R}^{\rm {H_{z}\times W_{z} \times C}}$ represent dynamic temporal template features.
The number of temporal templates stored in the collection is ${\rm N}$.
The dimensions are defined as follows: ${\rm H_{z}}$, ${\rm W_{z} = 6}$, ${\rm H_{x}}$, ${\rm W_{x} = 26}$, and the feature channel ${\rm C}$ is $256$.

To reduce the computation of the backbone and improve tracking speed, we store template features in the appearance collection instead of image representation to avoid repeatedly extracting features for templates.

\subsection{Mixed-Temporal Transformer}
The Mixed-Temporal Transformer (MTT) for feature fusion is built with an Encoder-Decoder architecture, as shown in Fig. \ref{fig:MTT}. 
MTT builds relationships between the search region and multiple target templates in parallel through a mixed-temporal attention mechanism, which is the critical module of our BACTrack.

\noindent\textbf{Transformer Encoder.}
The encoder of the MTT is responsible for encoding the target template features, which consists of a multi-head self-attention (MSA) block and a two-feed-forward-layer module.
In our framework, the transformer encoder simultaneously receives and encodes a set of template features $\mathbf{F}_{\rm z}^k$ in the template collection, including the initial frame ($k = 0$) and the subsequent ${\rm N}$ frames ($k=1, \dots, {\rm N}$).
For the computation of attention, we first flatten the input template features into $\mathbf{F}_{\rm z}^{k'}\in\mathbb{R}^{\rm {H_{z}W_{z}\times C}}$.
To enhance the perception of the target's position, we add a spatial position encoding $\mathbf{P}_{\rm z}^k\in\mathbb{R}^{\rm {H_{z}W_{z}\times C}}$ using a sine function\cite{vt2}. Other operations are the same as vanilla Transformer\cite{attention}.
The calculation of the encoder can be formulated as:
\begin{equation}
\hat{\mathbf{X}}_{\rm z}^k = {\rm Norm}({\rm MSA}(\mathbf{F}_{\rm z}^{k'}+\mathbf{P}_{\rm z}^k)+(\mathbf{F}_{\rm z}^{k'}+\mathbf{P}_{\rm z}^k)),
\end{equation}
\begin{equation} 
\mathbf{X}_{\rm z}^k = {\rm Norm}({\rm FFN}(\hat{\mathbf{X}}_{\rm z}^k)+\hat{\mathbf{X}}_{\rm z}^k).
\end{equation}
The transformer encoder generates a group of template-encoded features, which can be formulated as $\mathbf{X}_{\rm enc}=\left\{\mathbf{X}_{\rm z}^0,\mathbf{X}_{\rm z}^1,\dots,\mathbf{X}_{\rm z}^{\rm N}\right\}\in\mathbb{R}^{{\rm N\times H_{z}W_{z}\times C}}$.
Each template with a different appearance is encoded independently to obtain the enhanced template features and aggregated together for input to the decoder.

\begin{figure}
\centering
\includegraphics[width=1.0\linewidth]{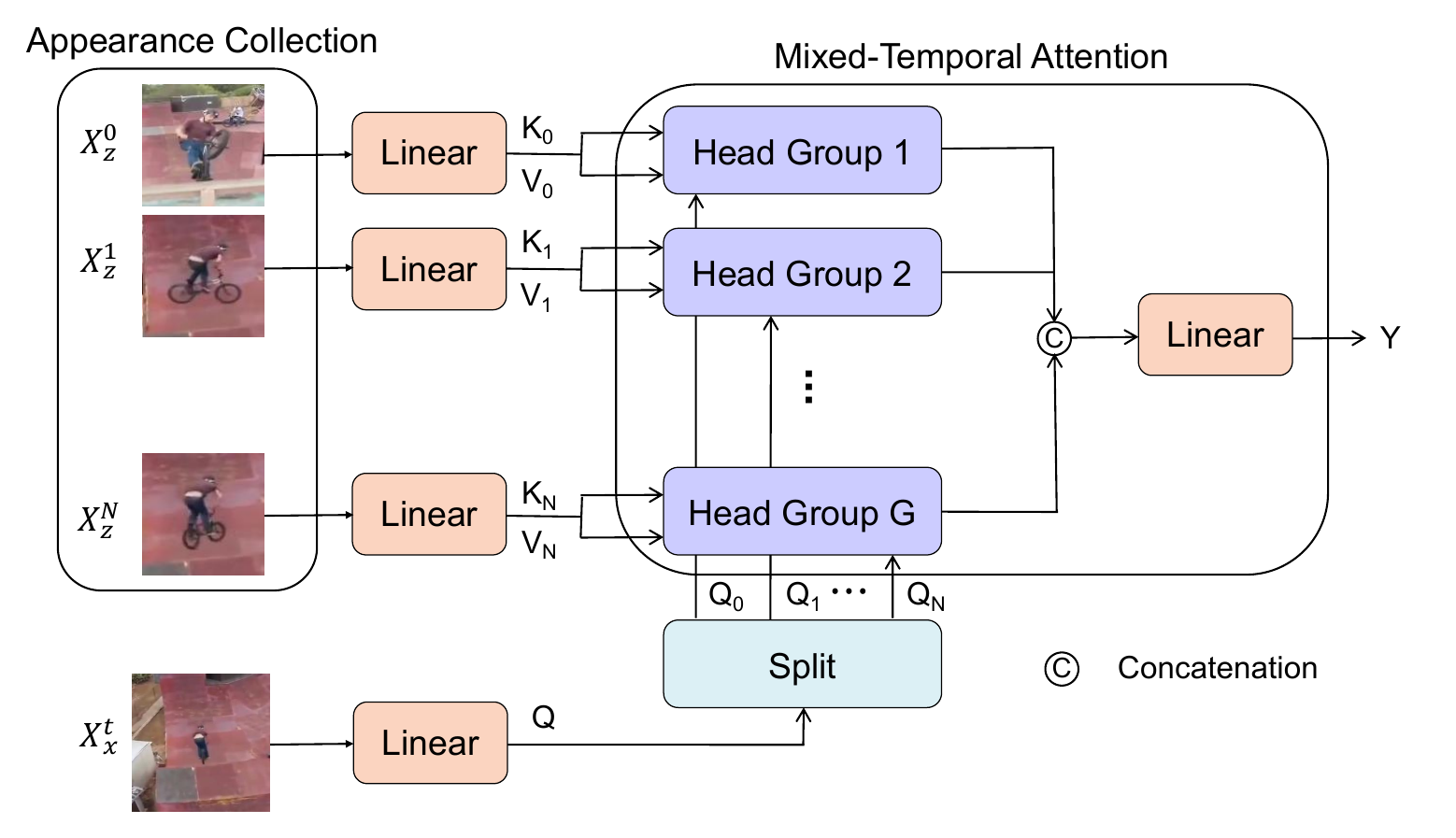}
\caption{\textbf{Architecture of the proposed Mixed-Temporal Attention (MTA).} MHA groups the head to achieve a fusion of the templates in the appearance collection with the current search feature.} 
\label{fig:attention}
\end{figure}

\noindent\textbf{Mixed-Temporal Attention.}
In order to fuse multiple templates without increasing computational demands, we propose a mixed-temporal attention mechanism (MTA), as depicted in Fig. \ref{fig:attention}, inspired by the multi-head attention (MHA)\cite{attention}. MHA simultaneously processes data through multiple attention mechanisms, each focusing on different parts of the input to capture a more diverse range of information.
Unlike the traditional MHA, where the keys and values are typically identical across all heads in a given attention layer, our MTA is designed to focus on different temporal information through different heads.
This unique structure allows each head to specialize in processing different temporal context, thereby enabling the MTA to capture a richer and more diverse range of temporal features without significantly adding to computational complexity.

\emph{Head grouping.} To perform attention calculation for multi-template feature fusion, we divide the heads into ${\rm G}$ groups, corresponding to the capacity of the appearance collection (\emph{i.e.}, ${\rm N}+1$).
This division allows each group to focus on different temporal information during the tracking process.
Specifically, the key $\mathbf{K}_{k}$ and value $\mathbf{V}_{k}$ in different group are obtained from linear projections of different templates.

\emph{Generating keys and values.}
Given the output of encoder $\mathbf{X}_{\rm enc}=\left\{\mathbf{X}_{\rm z}^0,\mathbf{X}_{\rm z}^1,\dots,\mathbf{X}_{\rm z}^{\rm N}\right\}$ and current source region feature $\mathbf{X}_{\rm x}^t\in\mathbb{R}^{\rm {H_{x}W_{x}\times C}}$, where $\mathbf{X}_{\rm z}^0$ is encoded initial template feature, $\left\{\mathbf{X}_{\rm z}^k\right\}_{k=1}^{{\rm N}}$ is the encoded temporal template features, and ${\rm N}$ is the number of temporal templates. We first get the group keys $\left\{\mathbf{K}_{k}\right\}_{k=0}^{{\rm N}}$ and group values $\left\{\mathbf{V}_{k}\right\}_{k=0}^{{\rm N}}$ as:
\begin{equation} 
\mathbf{K}_{k},\mathbf{V}_{k} = \mathbf{X}_{\rm z}^{k}\mathbf{W}^{\rm K}_k, \mathbf{X}_{\rm z}^{k}\mathbf{W}^{\rm V}_k ,  k = 0,1, \dots, {\rm N},
\end{equation}
where $\mathbf{W}^{\rm K}_k, \mathbf{W}^{\rm V}_k\in\mathbb{R}^{\rm C\times C/{\rm G}}$ are linear projections.

\emph{Generating queries.}
Based on the current source region features $\mathbf{X}_{\rm x}^t$, we also derive queries $\mathbf{Q}$ as:
\begin{equation}
 \mathbf{Q} = \mathbf{X}_{\rm x}^t\mathbf{W}^{\rm Q},
\end{equation}
where $\mathbf{W}^{\rm Q}\in\mathbb{R}^{\rm C\times C}$ is linear projection. To align with the head grouping, we also split the feature channels of the queries $\mathbf{Q}$ into ${\rm G}$ groups. The $k$-th channel group of $\mathbf{Q}$ denoted by $\mathbf{Q}_{k}=\mathbf{Q}\left [ :,k\times {\rm C}:(k+1)\times {\rm C} \right ], k = 0,1,\dots, {\rm N}$, is then fed into the corresponding head group for attention processing. 

\emph{Mixed-temporal attention calculation.}
Due to the grouping operation, each independent head group calculates cross-attention in parallel. 
The output of the $k$-th head group is
\begin{equation}
\mathbf{Y}_k = {\rm MHA}(\mathbf{Q}_k,\mathbf{K}_k,\mathbf{V}_k)={\rm Softmax}(\frac{\mathbf{Q} _k\mathbf{K}_k}{\sqrt{\rm d_k} } )\mathbf{V}_k,
\end{equation}
where ${\rm MHA}$(·) denotes standard multi-head attention, ${\rm \sqrt{d_k}}$ is a scaling factor.

We concatenate the output from all head groups and once again project to obtain the final mixed-temporal feature $\mathbf{Y}$:
\begin{equation}
\mathbf{Y} = {\rm Concat}(\mathbf{Y}_0, \dots, \mathbf{Y}_{\rm N}) \mathbf{W}^0, 
\end{equation}
where $\mathbf{W}^{0}$ is linear projection. Therefore, our mixed-temporal attention allows the model to jointly attend to information from different frames, resulting in multi-template aggregation.

\noindent\textbf{Transformer Decoder.}
The decoder of the Mixed-Temporal Transformer is a crucial component of BACTrack. It fuses the template features, which characterize different target appearances and are generated by the encoder, with the search feature through the mixed-temporal attention mechanism(MTA).
The input to the decoder includes a search feature $\mathbf{F}_{\rm x}^t$ with a spatial position encoding $\mathbf{P}_{\rm x}^t\in\mathbb{R}^{\rm H_{x}W_{x}\times C}$ and the collection of encoded template features $\mathbf{X}_{\rm enc}$.
Similar to the encoder, we reshape $\mathbf{F}_{\rm x}^t$ to $\mathbf{F}_{\rm x}^{t'}\in\mathbb{R}^{\rm H_{x}W_{x}\times C}$.
The reshaped search region feature $\mathbf{F}_{\rm x}^{t'}$ is then enhanced by MSA, which is the same as the template features:
\begin{equation}
\mathbf{X}_{\rm x}^t = {\rm Norm}({\rm MSA}(\mathbf{F}_{\rm x}^{t'}+\mathbf{P}_{\rm x}^t)+(\mathbf{F}_{\rm x}^{t'}+\mathbf{P}_{\rm x}^t)).
\end{equation}
Different from the decoder layer of vanilla Transformer\cite{attention}, for appearance collection, our decoder calculates cross-attention on multiple template features and current search features using MTA, which can be formulated as:
\begin{equation} 
\hat{\mathbf{X}}_{\rm dec} = {\rm Norm}({\rm MTA}(\mathbf{X}_{\rm x}^t,\mathbf{X}_{\rm enc},\mathbf{X}_{\rm enc})+\mathbf{X}_{\rm x}^t),
\end{equation}
\begin{equation} 
\mathbf{X}_{\rm dec} = {\rm Norm}({\rm FFN}(\hat{\mathbf{X}}_{\rm dec})+\hat{\mathbf{X}}_{\rm dec}).
\end{equation}
$\mathbf{X}_{\rm dec}$ denotes the mixed-temporal feature finally generated by MTT, which effectively associates the search region and mixed target templates.

Finally, the prediction head network of our framework consists of two branches: a classification head to predict the category at each location and a regression head to calculate the target's bounding box for that location, following \cite{xu2020siamfc++}.

\subsection{Appearance Discriminator
\label{sec:inference}}
We introduce an appearance discriminator (AD) to build a dynamic collection of target templates, which employs an online adaptive template-update strategy.
This strategy ensures the appearance collection of templates remains reliable and diverse, effectively adapting to rapid appearance changes of the target during tracking. 
The scale of appearance collection is fixed, including an initial template and ${\rm N}$ temporal templates, denoted as $\left\{\rm z_0, z_1, z_2,\dots, z_N\right\}$, where ${\rm z_0}$ is the initial template.
The update of our dynamic appearance collection follows the First In and First Out (FIFO) principle: when the appearance discriminator decides to update a temporal template, the current frame is added to the appearance collection, and simultaneously, the oldest reference frame in the cache is removed.

To address challenges in UAV tracking, our framework employs two critical judgmental principles: reliability and diversity. The reliability principle is prioritized, ensuring that only trustworthy templates, which accurately represent the target and exclude distractors or background elements, are retained in the collection.  
Following this, the diversity principle is applied to avoid excessive similarity among the stored templates, thereby preserving the advantages of our multi-temporal template fusion.

\noindent\textbf{Reliability Principle.}
For the reliability principle, the confidence score ${\rm s}_t$ generated by the classification head is considered an evaluation metric. 
Specifically, we apply a window penalty in the post-processing stage\cite{transT}.
A dynamic confidence score threshold $\tau_t$ is established, which is tailored to the specific tracking scenario and its progression. Specifically, the confidence scores from the initial five frames of each tracking video serve as a reference, allowing the threshold to adapt to diverse tracking scenarios.
When the count of updated templates is less than ${\rm N}$, the threshold is set at ${\rm w_1}$ times the mean of the confidence scores from the initial five frames. Conversely, if the number of updated templates reaches ${\rm N}$, the threshold is adjusted to ${\rm w_2}$ times the mean of the confidence scores within the appearance collection.
Initially, a constant threshold ${\rm \tau_{0}}$ is employed for the first five frames. For subsequent frames, the confidence score threshold for the $t$-th frame can be expressed as follows:
\begin{equation}
\tau_{t}=\left\{
\begin{aligned}
\tau_{0}&,             &t\le 5\\
{\rm w_1\cdot mean(s}_i)  &, &t>5, {\rm n}_{t} < {\rm N} \\
{\rm w_2\cdot mean(s}_j)  &, &t>5, {\rm n}_{t} \ge {\rm N} ,\\
\end{aligned}
\right.
\end{equation}
where ${\rm n}_{t}$ represents the count of updated templates, $i = 1,\dots,5$ represents the first five frames, and $j = 1,\dots,m$ corresponds to frames that have been updated within the appearance collection.

\noindent\textbf{Diversity Principle.}
Simultaneously, it is crucial to account for template diversity in our tracking process.
To address this, we employ the SSIM \cite{SSIM} as a measure of similarity. SSIM assesses structural similarity between two images, comprising three components: luminance, contrast, and structure. We compute the SSIM between the current frame image patch ${\rm z}_t$ and the most recent temporal template image patch in the appearance collection ${\rm z_N}$:
\begin{equation}
{\rm ssim}_t=\frac{(2\mu _{{\rm z}_t}\mu _{\rm z_N}+{\rm c_1})(\sigma_{{{\rm z}_t}{\rm z_N}}+{\rm c_2})}{(\mu_{{{\rm z}_t}}^2+\mu_{{\rm z_N}}^2+{\rm c_1} )(\sigma_{{{\rm z}_t}}^2+\sigma_{{\rm z_N}}^2+{\rm c_2})} ,
\end{equation}
where $\mu$ denotes average value, $\sigma_{{{\rm z}_t}}^2$ and $\sigma_{{\rm z_N}}^2$ denote the standard deviation of ${\rm z}_t$ and ${\rm z_N}$ respectively, $\sigma_{{{\rm z}_t}{\rm z_N}}$ denotes the covariance of ${\rm z}_t$ and ${\rm z_N}$, and $\rm c_1$, $\rm c_2$ are constants.
We evaluate templates within the collection using ${\rm ssim}_t$ and establish a threshold $\tau_{\rm si}$ as an updating criterion. 
If ${\rm ssim}_t$ surpasses $\tau_{\rm si}$, it implies that the current frame's outcome closely resembles the templates in the collection, thereby obviating the need for template updates.

To sum up, for the $t$-th frame, the updating and storing of template within the appearance collection must satisfy the following dual constraints:
\begin{equation}
\begin{aligned}
{\rm s}_{t}>\tau_t,\\
{\rm ssim}_{t}<\tau_{\rm si}.
\end{aligned}
\end{equation}
Here, ${\rm s}_{t}$ represents the confidence score of the $t$-th frame, ${\rm ssim}_{t}$ denotes the SSIM between the template of $t$-th frame and the most recent template in the collection. $\tau_{\rm si}$ is the parameter denoting the similarity threshold, while $\tau_t$ serves as the confidence score threshold, adapting to the specific tracking scenario and progression.

\begin{figure*}
\centering  
\vspace{-0.35cm}
\subfigtopskip=2pt 
\subfigbottomskip=2pt 
\subfigcapskip=-5pt
\subfigure{
\includegraphics[width=0.34\linewidth]{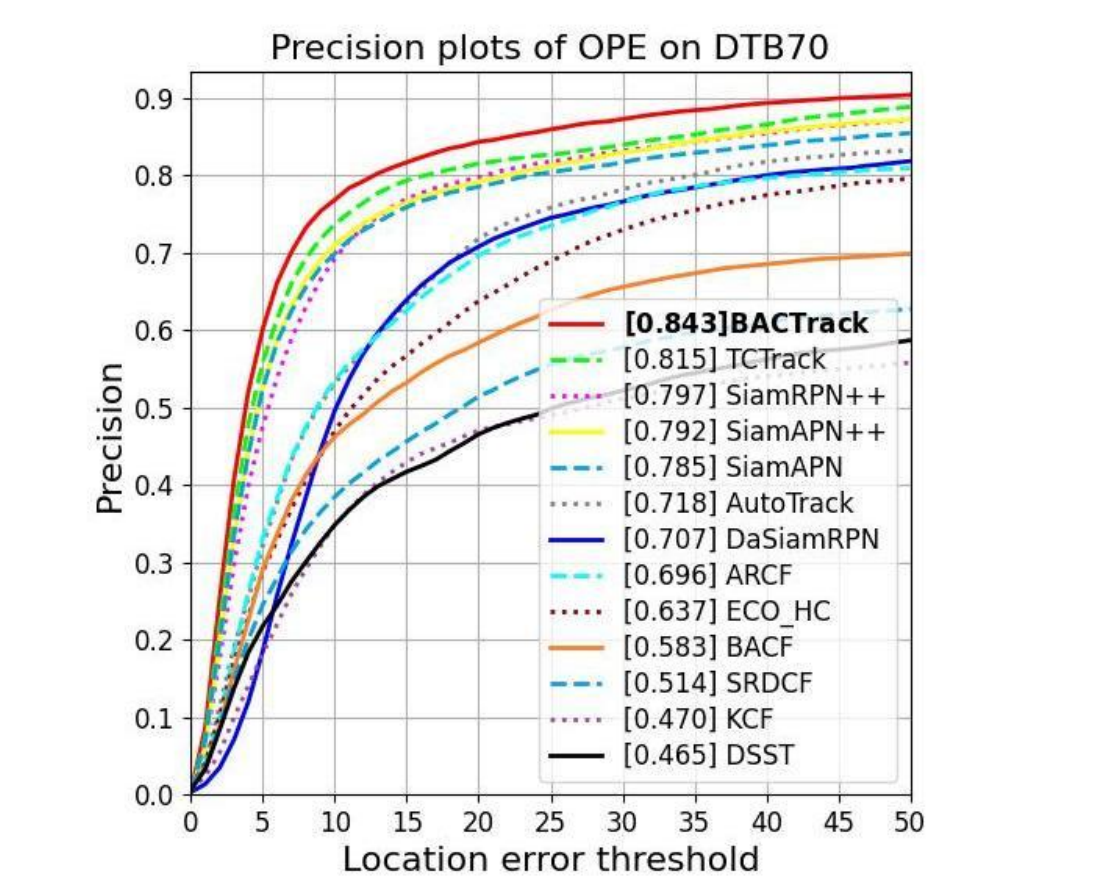}}\hspace{-10mm}
\subfigure{
\includegraphics[width=0.34\linewidth]{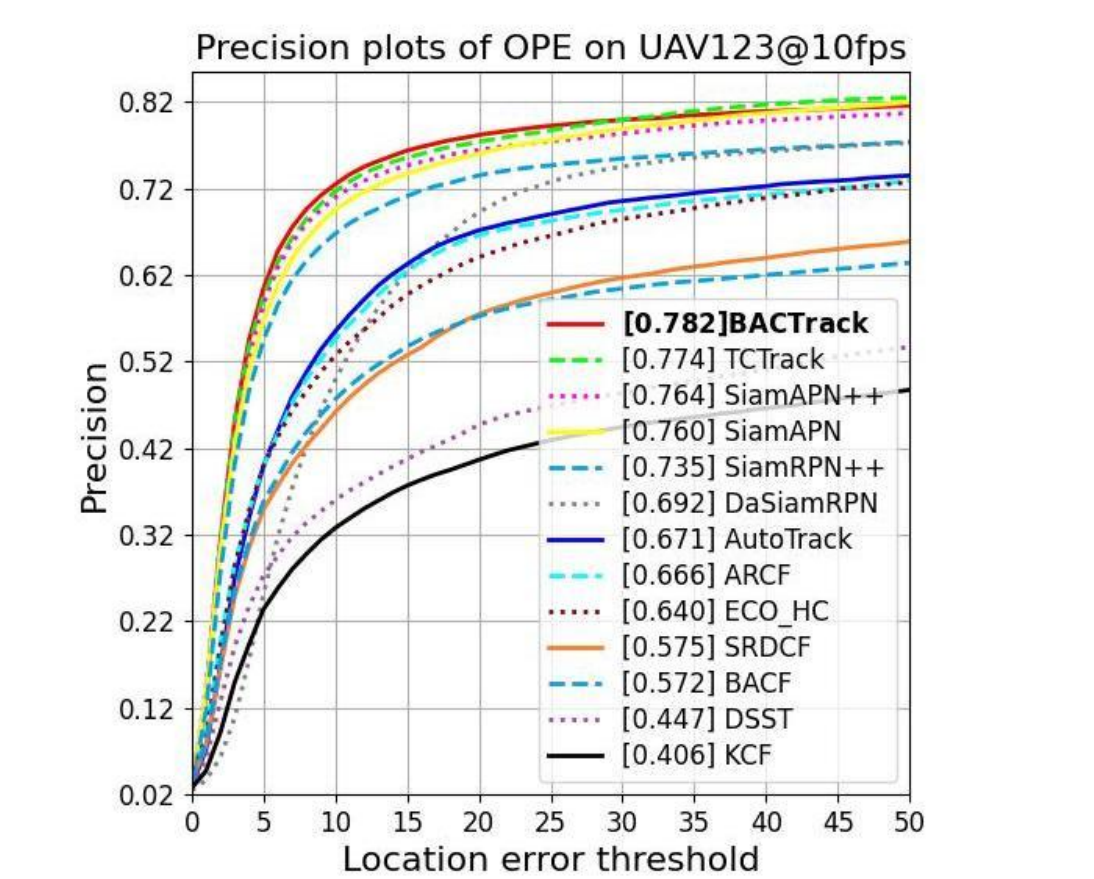}}\hspace{-10mm}
\subfigure{
\includegraphics[width=0.34\linewidth]{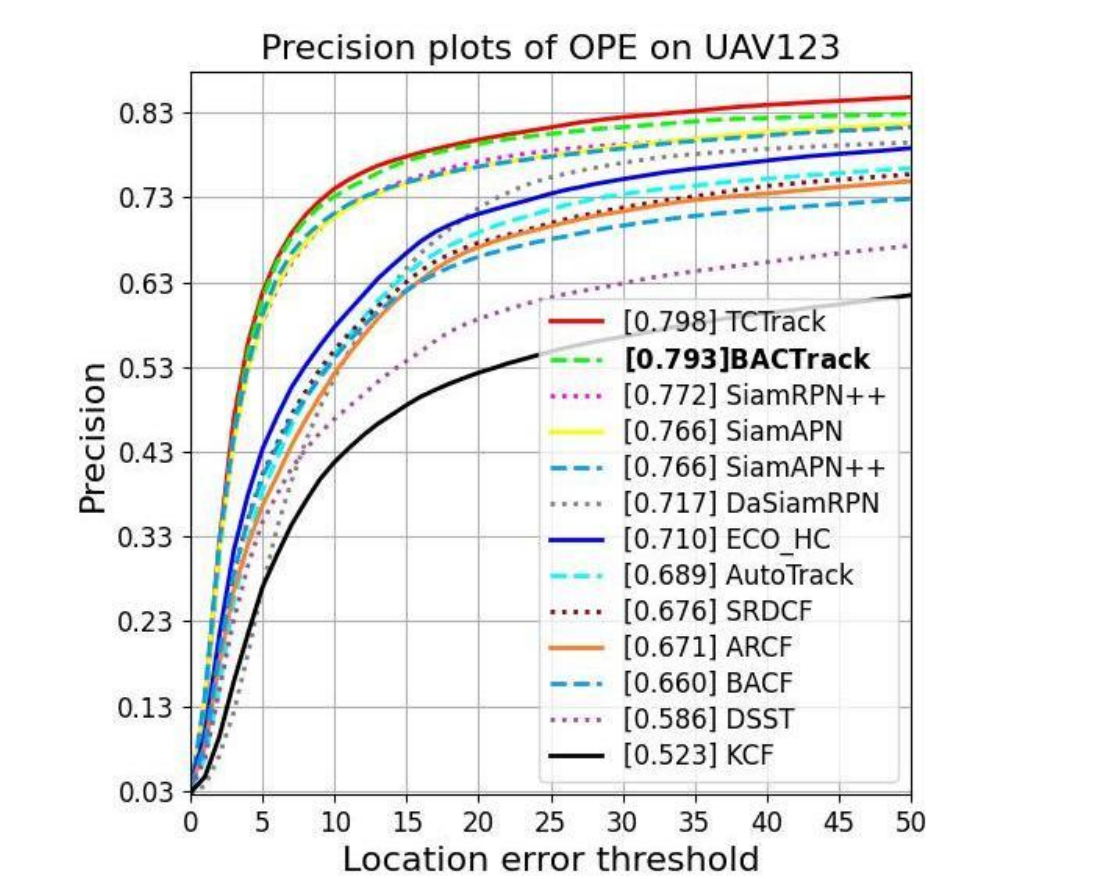}}\\
\subfigure{
\includegraphics[width=0.34\linewidth]{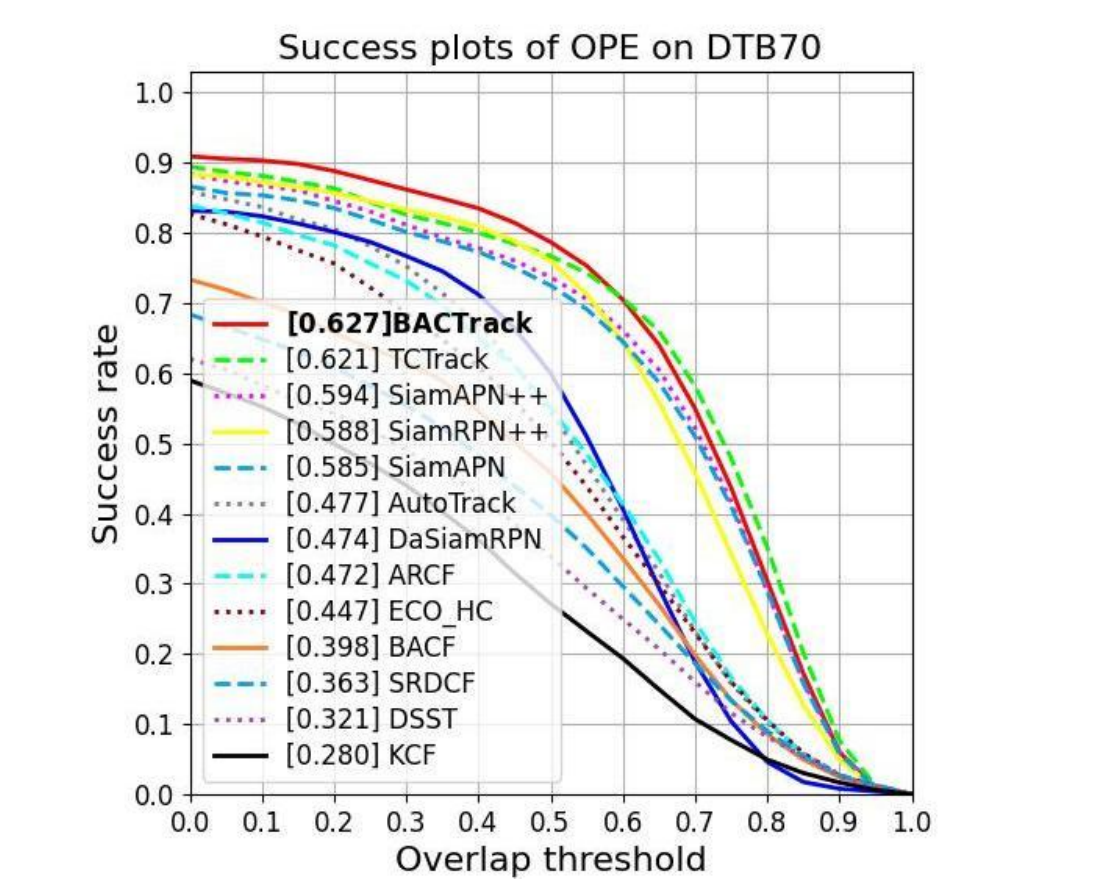}}\hspace{-10mm}
\subfigure{
\includegraphics[width=0.34\linewidth]{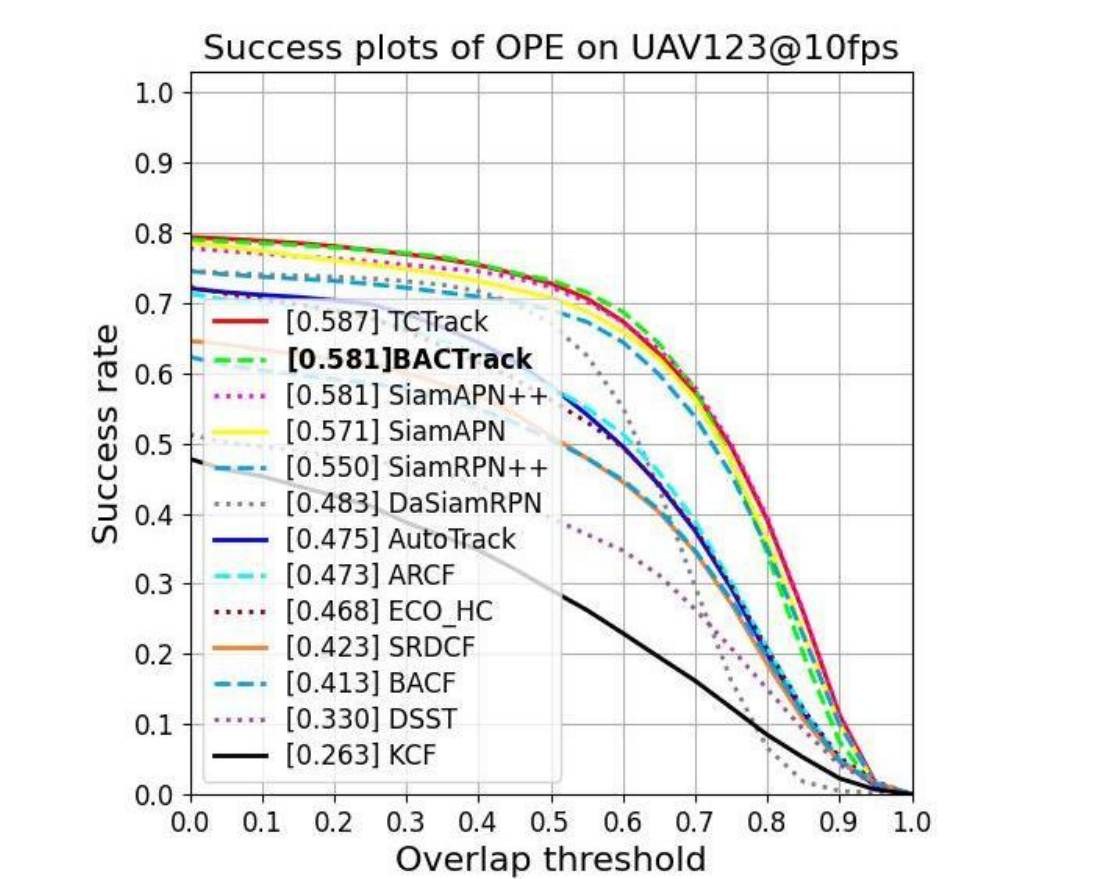}}\hspace{-10mm}
\subfigure{
\includegraphics[width=0.34\linewidth]{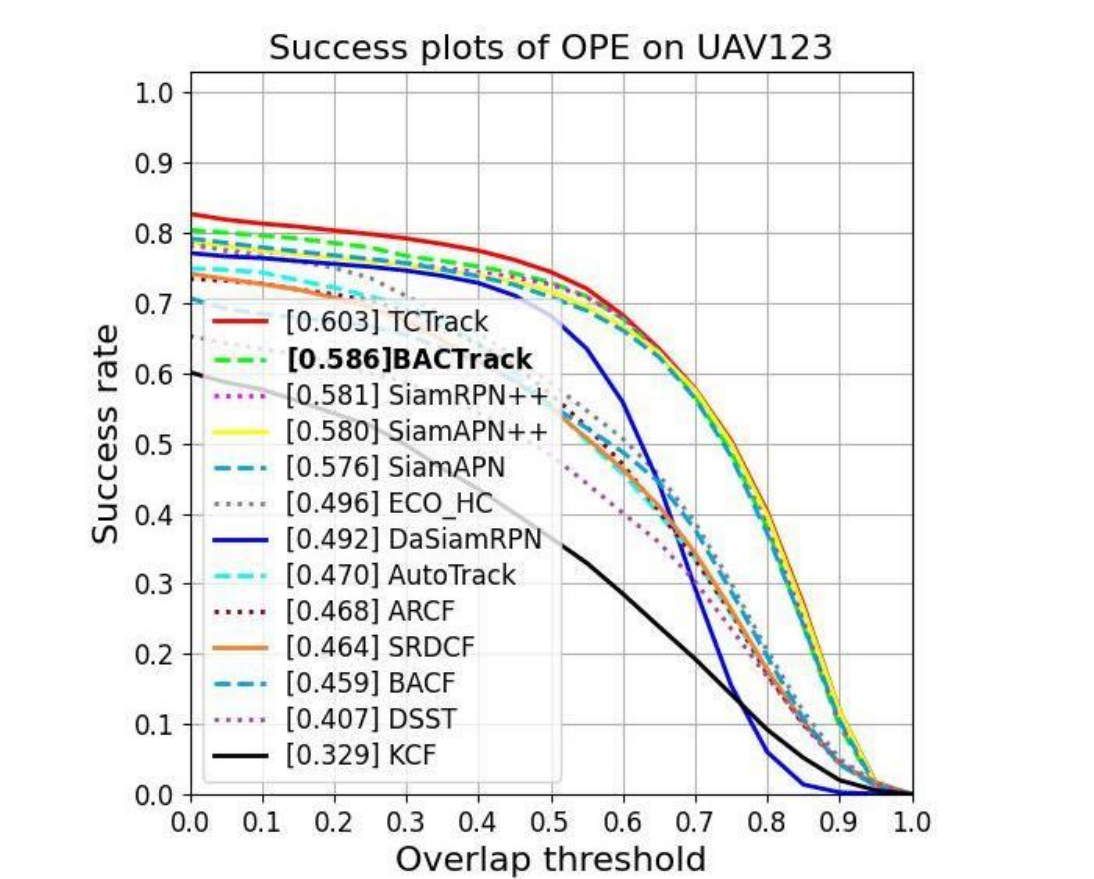}}
\caption{Overall performance of BACTrack and other state-of-the-art trackers on DTB70, UAV123@10fps, and UAV123.}
\label{fig:results}
\end{figure*}

\section{Experiments}
We comprehensively evaluate our proposed BACTrack on four authoritative aerial tracking benchmarks, \emph{i.e.}, DTB70\cite{DTB70}, UAV123\cite{uav123}, UAV123@10fps\cite{uav123}, and UAVTrack112\_L\cite{fu2021onboard112}. 
Existing superior trackers are included for a thorough comparison, including TCTrack\cite{TCTrack}, SiamRPN++\cite{siamrpn++}, DaSiamRPN \cite{DaSiamRPN}, SiamAPN++\cite{siamapn++}, SiamAPN\cite{SiamAPN}, SiameseFC\cite{SiamFC}, SiamFC++\cite{xu2020siamfc++} HIFT\cite{HiFT}, ECO\_HC\cite{eco},  AutoTrack\cite{AutoTrack}, ARCF\cite{ARCF}, SRDCF\cite{SRDCF}, BACF\cite{BACF}, DSST\cite{DSST}, and KCF\cite{KCF}, SiamMask\cite{siammask}, Ocean\cite{ocean}, SiamDW\_FC\cite{SiamDW}, SiamBAN\cite{siamban}, SiamCAR\cite{siamcar}, SiamGAT\cite{siamgat}, and SE\_SiamFC\cite{SESiamFC}.
Visualisation and ablation studies are further performed to confirm our performance and model design.

\subsection{Implementation Details}
\label{sec:details}
\noindent\textbf{Training Details.}
Our model is trained using the training splits from VID \cite{VID}, Lasot \cite{fan2019lasot}, and GOT-10K \cite{GOT-10k}, over 100 epochs on two GPUs. During the initial 10 epochs, the backbone parameters are frozen\cite{siamrpn++}. 
We employ a step learning rate scheduler, initiating with a warmup period of 5 epochs, where the learning rate ascends logarithmically from $0.005$ to $0.01$. Subsequently, the learning rate logarithmically decreases from $0.01$ to $0.00005$ for the rest of the training period. The network is optimized using stochastic gradient descent (SGD) with a momentum of 0.9, utilizing mini-batches comprising 100 pairs.

For training, we select two frames from each video with a maximum frame index difference of $100$ to serve as the the template and the search region. Additionally, three random frames preceding the search region are chosen as dynamic templates.
The input sizes of the template and the search images are set at $127\times 127$ and $287\times 287$ respectively.  
In MTA, we use 8 heads, divided into 4 groups (G), with 2 heads allocated to each group for keys and values. Each branch of the prediction head is composed of 5 Conv-BN-ReLU layers.

\noindent\textbf{Inference Details.}
During inference, we configure $\tau_{0}$, ${\rm w}_1$, ${\rm w}_2$, and $\tau_{\rm si}$ as $1.8$, $0.95$, $0.9$, and $0.42$, respectively.
The template collection consists of an initial template and three temporal templates.
To assess the efficiency and viability of BACTrack, we conducted a comparative analysis of tracking speed and precision against SOTA trackers. The testing process on PC was implemented on NVIDIA GeForce RTX 3090 Ti. The real-world testing was performed on an NVIDIA Jetson AGX Xavier.

\vspace{-0.1cm} 
\subsection{Quantitative Results
\label{sec:res}}
We comprehensively compare our tracker with existing efficient trackers on four standard aerial tracking benchmarks.
For Siamese trackers, we evaluate them using a lightweight backbone, AlexNet. 
The test codes and results for certain trackers were sourced from \cite{fu2021correlation} and \cite{fu2023siamese}. All evaluations were conducted using the One-Pass Evaluation (OPE) tracking method, with performance measured in terms of precision (Pre.) and success rate (Suc.).

\noindent\textbf{Results on DTB70.} DTB70 \cite{DTB70} is a diverse benchmark video dataset comprising 70 videos captured by drones, encompassing various UAV movements, including rapid translation and rotation.
As illustrated in Fig. \ref{fig:results}, our BACTrack achieves an AUC of $62.7\%$ and a precision of $84.3\%$, surpassing the baseline tracker, TCTrack, by $1\%$ in AUC and $3.4\%$ in precision. This achievement places BACTrack ahead of all previously published lightweight trackers. We attribute this success to BACTrack's template-update strategy, which proves especially effective in handling the numerous viewpoint changes in DTB70, where the appearance collection plays a pivotal role.

\begin{figure*}[ht]
\centering
\includegraphics[width=0.9\linewidth]{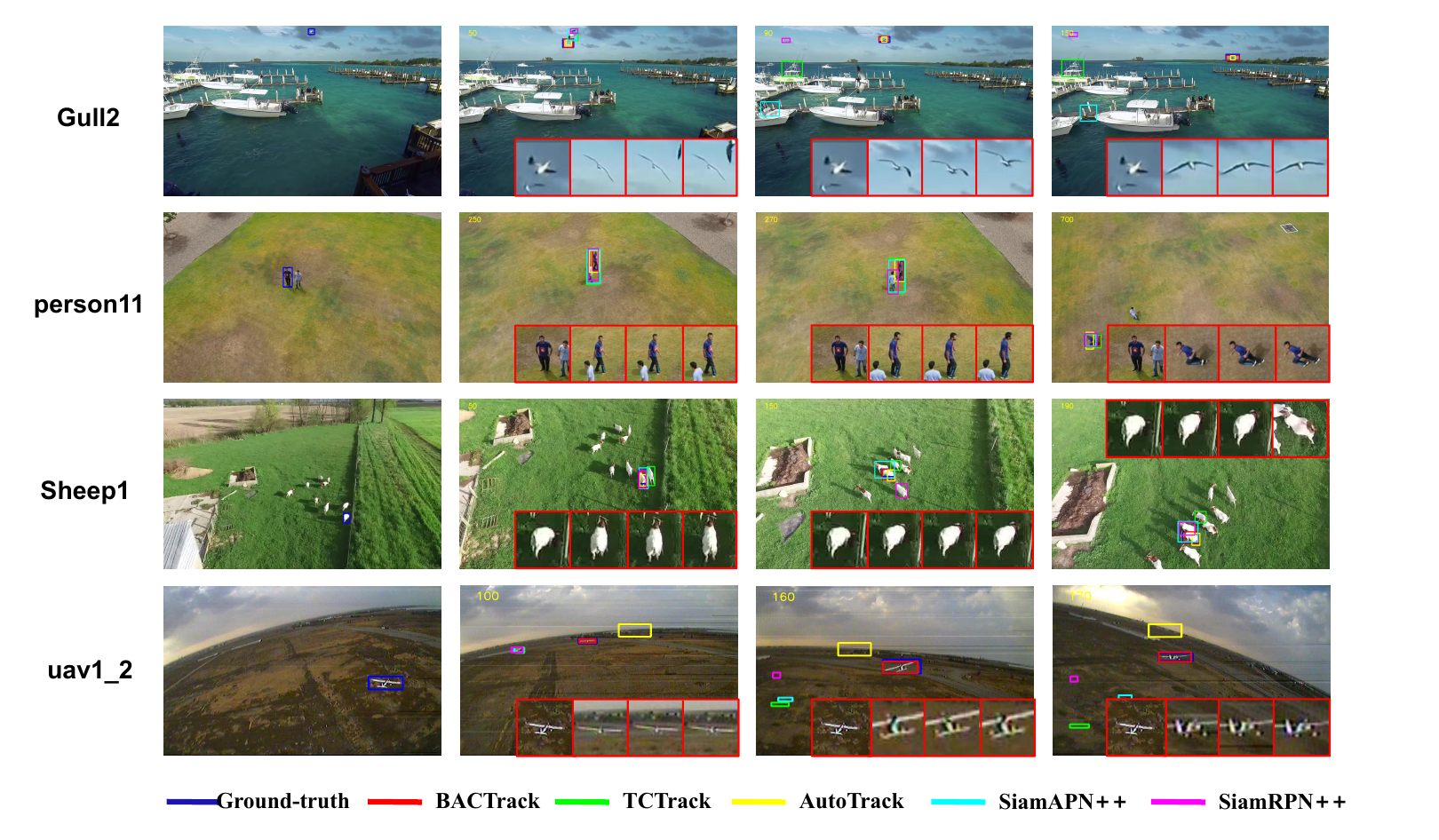}
\vspace{-3mm}
\caption{\textbf{Qualitative comparison of BACTrack with other top-performing trackers on some challenging sequences.} The ground-truth bounding boxes and the trackers' predicted boxes are colored differently. The appearance collection in the corner is marked with red boxes.}
\label{fig:tracking}
\end{figure*}
\begin{table*}[ht] 
\renewcommand\arraystretch{1.2}
\scriptsize 
\tabcolsep=0.9mm
\caption{\textbf{Overall performance on UAVTrack112\_L.} The best performance is marked in \textbf{bold}.}
\centering
\setlength{\tabcolsep}{0.5mm}{
\begin{tabular}{c|c c c c c c c c c c| c}
\toprule
Trackers & 
SRDCF\cite{SRDCF}&
ARCF \cite{ARCF} & 
BACF \cite{BACF} &
AutoTrack\cite{AutoTrack} & 
SiameseFC\cite{SiamFC} & 
DaSiamRPN\cite{DaSiamRPN} & 
HiFT\cite{HiFT} & 
SiamAPN++\cite{siamapn++} & 
SiamRPN++\cite{siamrpn++} & 
TCTrack\cite{TCTrack} &
\textbf{BACTrack}\\
\midrule
Suc.& 0.320 & 0.399 & 0.358    & 0.405 & 0.452 & 0.479 & 0.551 & 0.537 &0.559 &0.582 & \textbf{0.601 }\\
Pre.& 0.508 & 0.640 & 0.593  & 0.675 & 0.690 & 0.729 & 0.734  & 0.735 & 0.773& 0.786 & \textbf{0.787} \\
\bottomrule
\end{tabular}}
\label{tab:112}
\vspace{-3mm}
\end{table*}

\begin{table*}[ht] 
\renewcommand\arraystretch{1.2}
\scriptsize 
\caption{\textbf{Average evaluation on four aerial tracking benchmarks.} The best performance is marked in \textbf{bold}.}
\tabcolsep=0.9mm
\centering
\setlength{\tabcolsep}{0.5mm}{
\begin{tabular}{c|c c c c c c c c c c |c}
\toprule
Trackers & 
SRDCF\cite{SRDCF}&
KCF\cite{KCF} &
BACF\cite{BACF}&
ARCF\cite{ARCF} &
AutoTrack\cite{AutoTrack} & 
DaSiamRPN\cite{DaSiamRPN}& 
SiamRPN++\cite{siamrpn++} & 
SiamAPN\cite{SiamAPN} & 
SiamAPN++\cite{siamapn++} & 
TCTrack\cite{TCTrack}& 
\textbf{BACTrack}\\
\midrule
Suc. & 0.393 & 0.289 & 0.407& 0.453 & 0.457 & 0.482 & 0.570 & 0.586 & 0.573 & 0.598& \textbf{0.599} \\
Pre. & 0.568  & 0.486 & 0.602 & 0.669 & 0.689 & 0.711 & 0.769 &0.780 & 0.765 & 0.794 & \textbf{0.801}\\
\bottomrule
\end{tabular}}
\label{tab:avevs}
\vspace{-3mm}
\end{table*}
\noindent\textbf{Results on UAV123.} UAV123 \cite{uav123} encompasses 123 video sequences captured from low-altitude unmanned aerial vehicles, comprising over 110,000 frames. Many sequences within this dataset exhibit characteristics typical of aerial images, including low-resolution motion blur.
As depicted in Fig. \ref{fig:results}, our approach attains an AUC of $58.6\%$ and a precision of $79.3\%$, ranking second in success rate and precision. This underscores the effectiveness of BACTrack in aerial tracking scenarios, specifically highlighting the advantageous impact of the mixed-temporal transformer's fusion of multiple temporal templates.

\noindent\textbf{Results on UAV123@10fps.} UAV123@10fps \cite{uav123} consists of 123 sequences, temporally down-sampled to 10 FPS, presenting formidable target displacements between frames and thereby elevating the complexity of tracking tasks. 
Despite these challenges, as illustrated in Fig. \ref{fig:results}, our method demonstrates superior robustness and accuracy, achieving a precision score of 0.782 and an AUC of 0.581. Notably, the precision of our BACTrack surpasses all other SOTA trackers. This highlights BACTrack's capacity to effectively handle abrupt changes in target appearance and significant positional shifts between adjacent frames, typical challenges encountered in UAV tracking.

\noindent\textbf{Results on UAVTrack112\_L.} UAVTrack112\_L\cite{fu2021onboard112} contains 112 sequences, including over ${\rm 60K}$ frames, thereby constituting the largest long-term aerial tracking benchmark to date.
As delineated in Table \ref{tab:112}, a comparative analysis with other leading trackers highlights BACTrack's enhanced robustness and accuracy, evidenced by its precision ($0.787$) and success rate ($0.601$). Notably, the AUC score surpasses the previous best tracker by $3\%$. Such exceptional performance on UAVTrack112\_L underscores BACTrack's efficacy in long-term tracking scenarios. By dynamically updating the target's appearance within temporal templates, our tracker ensures sustained robustness and reliability throughout extended tracking periods.

\noindent\textbf{Average Performance.} 
Table \ref{tab:avevs} reports the average precision and success rate of the 10 lightweight trackers and BACTrack on the four benchmarks. BACTrack achieves the highest average success rate score of 0.599 and precision score of 0.801. 
Remarkably, BACTrack outperforms the second-best tracker, TCTrack (0.794), by $0.8\%$ in precision. To our knowledge, BACTrack stands as the premier lightweight tracker for aerial tracking across the four benchmarks, excelling in both success rate and precision.

\subsection{Qualitative Results}
\noindent\textbf{Visualization of Tracking Results.}
To illustrate the effectiveness of our tracker, Fig. \ref{fig:tracking} provides qualitative comparisons between BACTrack against other top-performing trackers across challenging sequences. 
Four screenshots are selected for each sequence, including the initial frame, along with the current appearance collection of BACTrack.
BACTrack shows its superiority over other UAV trackers in handling challenging scenarios, including deformation (the first sequence), viewpoint change (the second sequence), similar objects (the third sequence), background clutter (the last sequence), among others.
The integration of temporal templates from the appearance collection is pivotal in achieving BACTrack's superior tracking performance.

\noindent\textbf{Visualization of Template Selection.}
\begin{figure}
\centering
\includegraphics[width=1.0\linewidth]{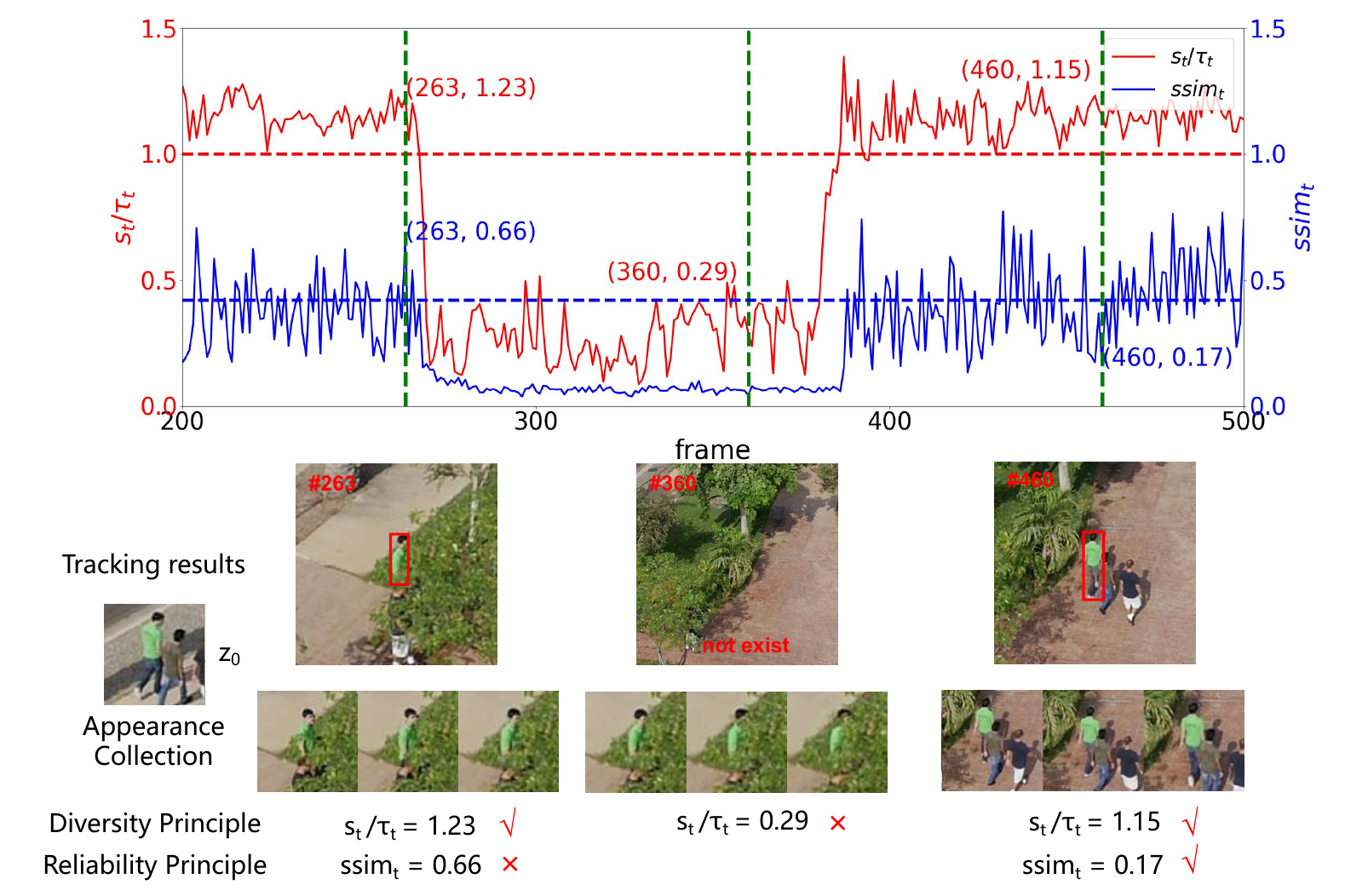}
\setlength{\abovecaptionskip}{-0.3cm} 
\caption{\textbf{Visualization of our template-update strategy.} Templates will be updated only if the reliability principle and diversity principle are both satisfied.}
\label{fig:update}
\vspace{-3mm}
\end{figure}
The template-update strategy of the appearance discriminator comprises two parts: the reliability principle and the diversity principle.
Fig. \ref{fig:update} provides a visual representation of our template-update strategy.
During the tracking process, each frame's tracking result is input into the appearance discriminator.
In the discriminator, the primary principle under consideration is reliability. Consequently, only tracking results with confidence scores (${s}_t$) exceeding the threshold ($\tau_{t}$) are considered for updates. This selective approach effectively filters out disruptive templates arising from occlusions or tracking errors (\emph{e.g.}, frame 360), ensuring the purity of the appearance collection and reducing the impact of background interference in aerial scenarios.
While the diversity principle is responsible for filtering out templates with high similarity to the appearance collection to maintain cosmetic diversity. Templates closely resembling the appearance collection are discarded (\emph{e.g.}, frame 263). 

Tracking results that conform to both of these principles are included into the appearance collection (\emph{e.g.}, frame 460), ensuring that the templates are both accurate and adaptable.
Thanks to the effectiveness of our template-update strategy, BACTrack excels at perceiving the target under its current unknown appearance within complex backgrounds.

\begin{figure*}
\centering
\includegraphics[width=0.95\linewidth]{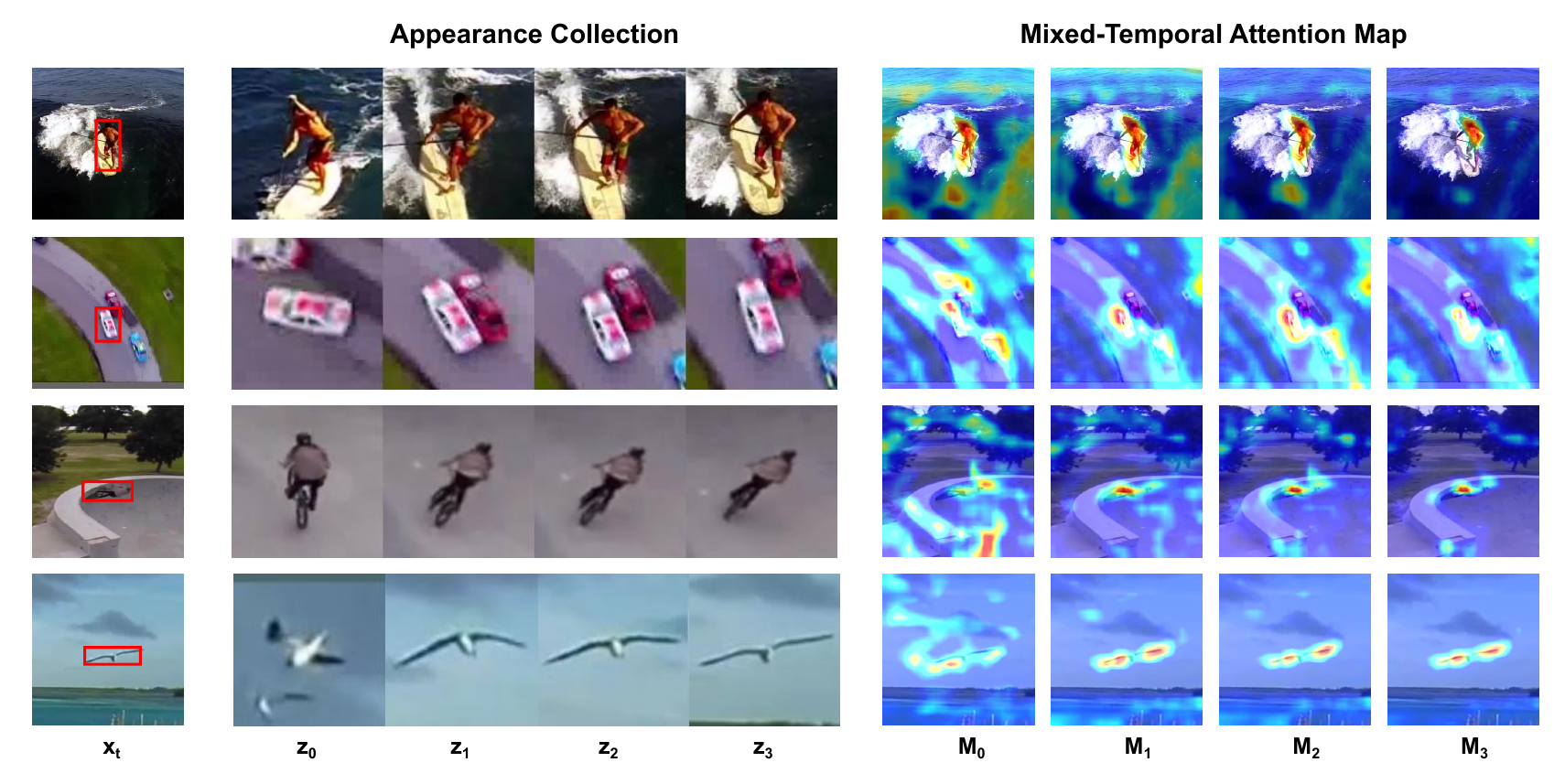}
\vspace{-3mm}
\caption{\textbf{Visualization of collected templates and corresponding attention maps.} 
Target objects are highlighted with red boxes. ${\rm z_0}$, ${\rm z_1}$, ${\rm z_2}$, and ${\rm z_3}$ denote the templates in the appearance collection. $\rm M_0$, $\rm M_1$, $\rm M_2$, and $\rm M_3$ represent the visualizations of the attention maps corresponding to the appearance collection for each head group in the MTA. Here, $\rm M_0$ is the attention map computed by the head group associated with the initial template, while the others pertain to temporal templates.}
\label{fig:vis}
\end{figure*}

\noindent\textbf{Visualization of Attention Maps.}
To validate the efficacy of mixed-temporal attention (MTA) in our BACTrack, Fig.~\ref{fig:vis} presents a visualization of the attention map.
The left column displays cropped images of the current search region (${\rm x}_t$). 
In the middle column, the appearance collection of templates for the current frame is depicted, comprising the initial template (${\rm z_0}$) and three temporal templates (${\rm z_1}$, ${\rm z_2}$, ${\rm z_3}$). 
The right column of the figure displays visualizations of the attention map generated by each head group in the MTA.
The first column of the attention map ($\rm M_0$) is computed by the head group associated with the initial template, while the subsequent columns represent the attention maps for the corresponding temporal templates ($\rm M_1$, $\rm M_2$, $\rm M_3$).

As shown in the attention map, $\rm M_0$ may pay attention to distractors or cluttered backgrounds.
In contrast, head groups associated with the temporal templates serve to rectify this issue.
Given the frequent substantial divergence of the target's appearance from the initial template, the employment of temporal templates is crucial for achieving accurate tracking.
The introduction of the appearance collection serves as a well-balanced complement to the initial template, providing diversity and a closer resemblance to the altered target's appearance. Therefore, the incorporation of appearance collection enhances the tracker's ability to accurately track target deformations.

\begin{figure}
\centering
\includegraphics[width=1\linewidth]{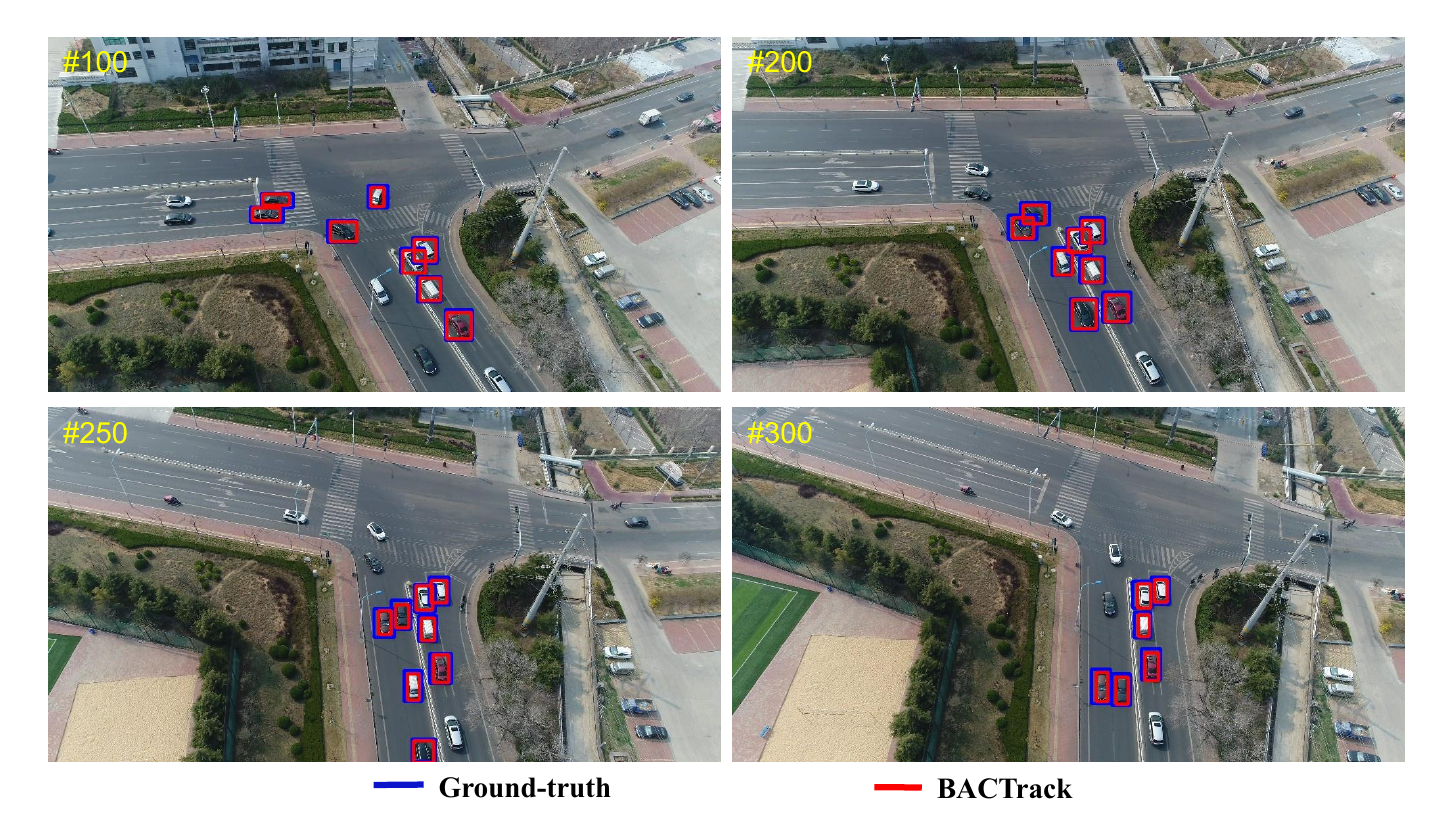}
\vspace{-3mm}
\caption{\textbf{Tracking performance in multiple similar objects scenes.} BACTrack can accurately track every single target in complex environments.}
\label{fig:uavdt}
\end{figure}

\noindent\textbf{Visualization of Tracking Multiple Similar Targets.}
To assess the performance of our tracker in aerial scenarios with a large number of visually similar objects, we conduct extended experiments on a multiple object tracking dataset. UAVDT\cite{uavdt} is a dataset focusing on vehicle traffic content captured by UAV. These tests aim to validate our tracker's ability to distinguish and accurately track individual object amidst multiple interfering targets.
We select a video sequence, M0101, from the UAV-benchmark-M, known for its complexity with multiple similar targets. In this sequence, we track each of eight targets in the sequence separately. Each object is given a ground-truth box at its first appearance. The result is shown in the Fig. \ref{fig:uavdt}. In scenarios where multiple targets of the same type are present, BACTrack accurately tracks each target without confusion from interfering objects. These extended experiments on the UAVDT demonstrate the ability of BACTrack to build an accurate appearance collection for each specified target, effectively dealing with similar objects.

\subsection{Efficiency Analysis
\label{sec:speed}}
\begin{table}[t]
\renewcommand\arraystretch{1.3}
\centering
\caption{\textbf{Comparison of FLOPs, parameters, inference speed (FPS), and performance on DTB70 with SOTA trackers.} We categorize all trackers into two groups: Trackers with deep backbones (Deep) and trackers with lightweight backbones (Light). }
\setlength{\tabcolsep}{1.5mm}{
\begin{tabular}{c|c|ccc|cc}
\toprule
& Trackers      &  \makebox[0.04\textwidth][c]{FLOPs} &  \makebox[0.04\textwidth][c]{\#Params} & \makebox[0.04\textwidth][c]{Speed}  &\makebox[0.04\textwidth][c]{Suc.} &\makebox[0.04\textwidth][c]{Pre.}\\ \midrule
\multirow{7}{*}{Deep} &  SiamRPN++\cite{siamrpn++} & 48.9 G     & 54.0 M     & 51.6  & 0.615&0.800    \\
&SiamMask\cite{siammask}  & 16.7 G     & 18.8 M      & 53.8   &  0.575& 0.777 \\
&Ocean\cite{ocean}      &4.4 G     & 25.9 M  & 61.6  &0.455 & 0.636     \\
&SiamDW\_FC\cite{SiamDW}     & 8.0 G     & 7.4 M       & 76.7  &0.489&0.722     \\
&SiamBAN\cite{siamban}   & 48.9 G     & 53.9 M      & 53.0 &  \textbf{0.643}  &0.834   \\
&SiamGAT\cite{siamgat}    & 17.3 G    & 14.2 M      & 46.4  &0.582&0.753     \\
&SiamCAR\cite{siamcar}   & 48.7 G    & 51.4 M      & 47.4  &0.603 &0.833     \\ \midrule
\multirow{5}{*}{Light}&  DaSiamRPN\cite{DaSiamRPN}             & 21.1 G    & 19.6 M      & \textbf{163.2}  &0.474 &0.707    \\
&SiamAPN\cite{SiamAPN}              & 10.3 G    & 15.1 M      & 127.5  &0.585&0.785    \\
&SiamAPN++\cite{siamapn++}            & 9.2 G     & 12.2 M      & 93.4 &0.594&0.792      \\
&TCTrack\cite{TCTrack}  & \textbf{4.51 G}     &\textbf{6.26 M}      & 76.4  &0.620 &0.814     \\
\cline{2-7}
&\textbf{BACTrack (Ours)}       & 5.04 G   & 6.52 M     & 87.2 &   0.627&\textbf{0.843}   \\ \bottomrule
\specialrule{0em}{1pt}{1pt}
\end{tabular}}
\label{tab:para}
\vspace{-3mm}
\end{table}
To align with the computational resource constraints typical of UAV platforms, our tracker is designed to be lightweight without compromising on accuracy.
Table \ref{tab:para} shows an efficiency analysis comparing BACTrack with other SOTA trackers, covering metrics such as FLOPs, parameters, inference speed, and tracking performance on the DTB70.
For an intuitive comparison, we categorize all trackers into two groups: those with deep backbones (on the top) and lightweight backbones (on the bottom).
The FLOPS ($\rm 5.04 G$) and Parameters ($\rm6.52 M$) of BACTrack surpass those of all trackers with deep backbones and outperform most lightweight trackers.
In terms of speed analysis, our BACTrack achieves a real-time speed of $87.2$ FPS on NVIDIA GeForce RTX 3090 Ti, surpassing all trackers with deep backbones.
Among the lightweight trackers, BACTrack attains the highest success rate ($0.627$) and precision ($0.843$). Additionally, it outperforms TCTrack\cite{TCTrack} in speed, exemplifying BACTrack's efficiency and effectiveness.

The remarkable model complexity and tracking speed of BACTrack suggest that the introduction of multiple temporal templates through MTA improves the accuracy and robustness of the model, all while conserving computational resources.
\begin{figure}[t]
\centering
\includegraphics[width=1.0\linewidth]{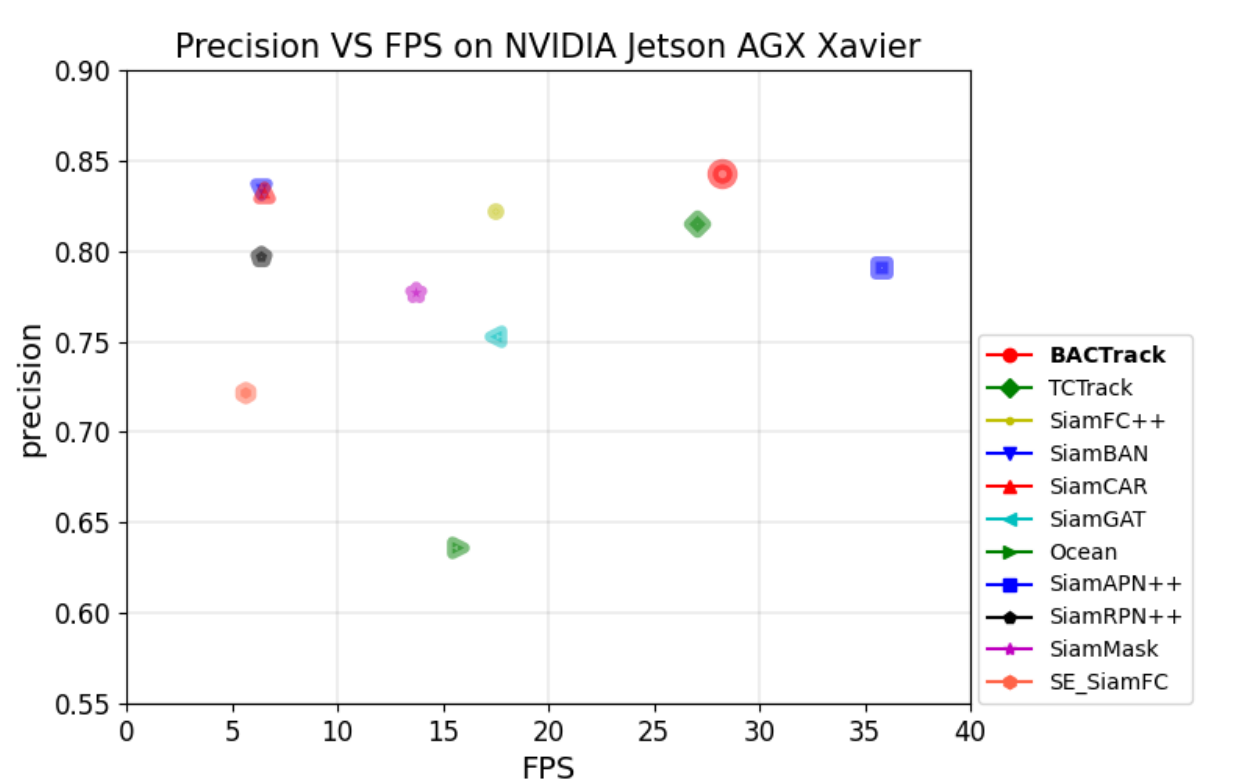}
\caption{\textbf{Performance comparison of BACTrack and sota trackers deployed on NVIDIA Jetson AGX Xavier.} Our tracker realizes the trade-off between speed and accuracy.}
\label{fig:agx}
\vspace{-3mm}
\end{figure}
\subsection{Embedded Platform Tests}
Deploying aerial tracking algorithms on resource-constrained drone hardware has consistently presented a formidable challenge. Many existing drone platforms suffer from limited computational power, leading to compromises in algorithm performance and real-time responsiveness. 
Therefore, to validate the applicability of our tracker, we further conduct real-world tests by deploying BACTrack on NVIDIA Jetson AGX Xavier, an embedded computing board designed for edge AI applications. This board is well-suited for UAV applications due to its compact form factor, energy efficiency, and high computational capabilities.
Fig. \ref{fig:agx} shows the performance comparison between our tracker and other excellent trackers deployed on the embedded platform. 
BACTrack achieves a speed of over $28$ FPS with high precision, balancing accuracy and speed. Our results indicate that BACTrack is not only highly effective but also practical for real-time applications in UAV-based scenarios.

\begin{figure*}
\centering  
\vspace{-0.35cm} 
\subfigtopskip=2pt 
\subfigbottomskip=2pt 
\subfigcapskip=-5pt 
\subfigure{
\includegraphics[width=0.3\linewidth]{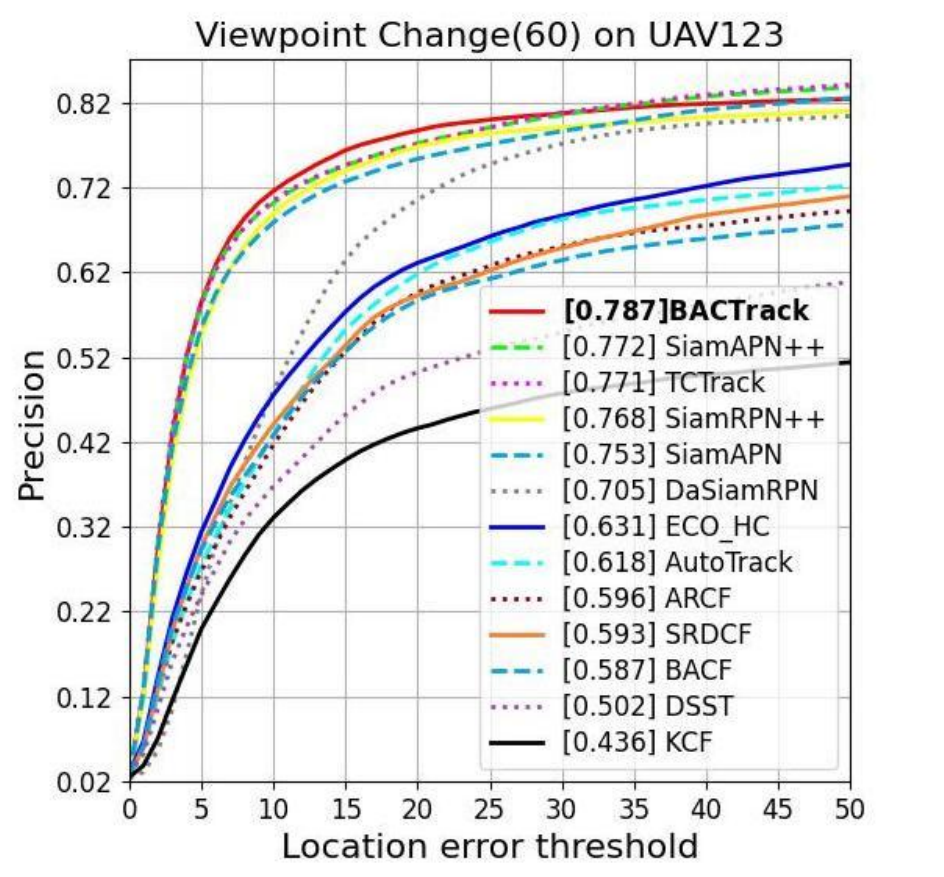}}\hspace{-3mm}
\subfigure{
\includegraphics[width=0.3\linewidth]{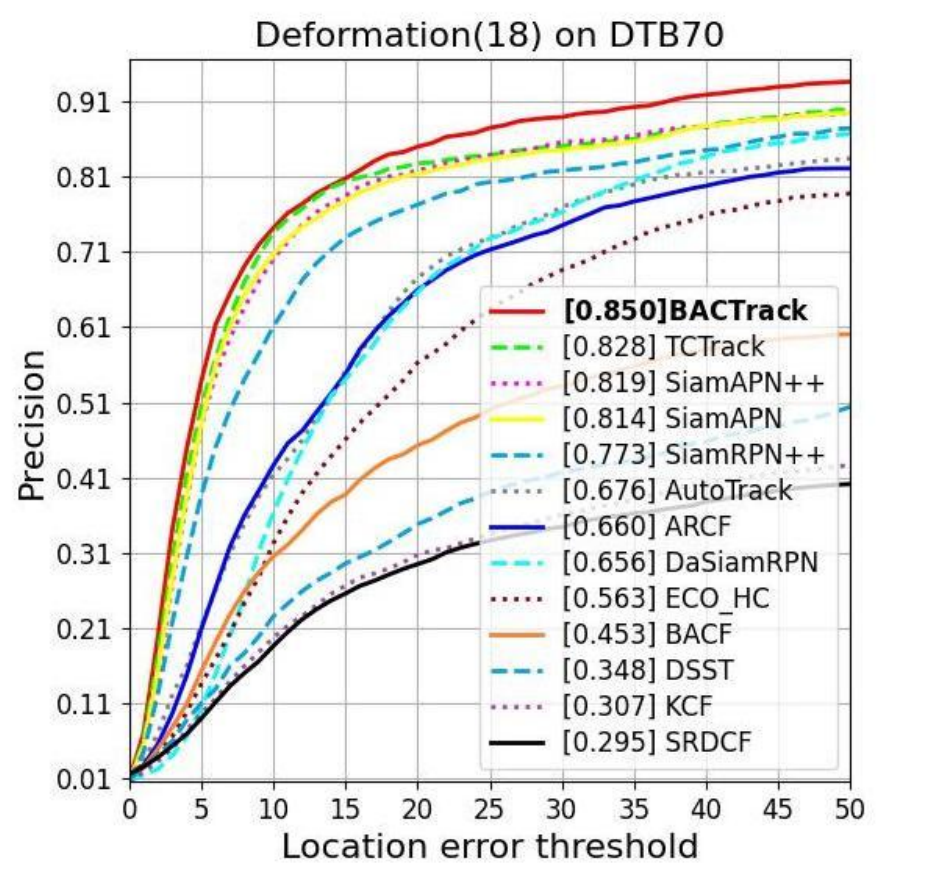}}\hspace{-3mm}
\subfigure{
\includegraphics[width=0.3\linewidth]{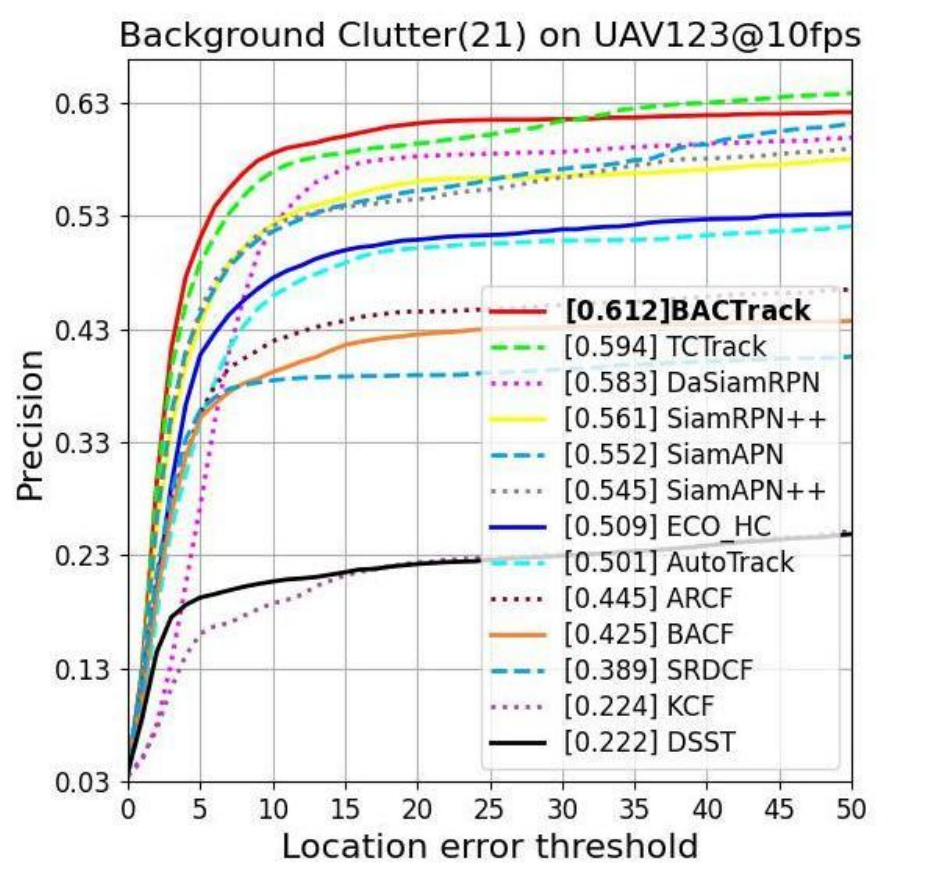}}\\
\subfigure{
\includegraphics[width=0.3\linewidth]{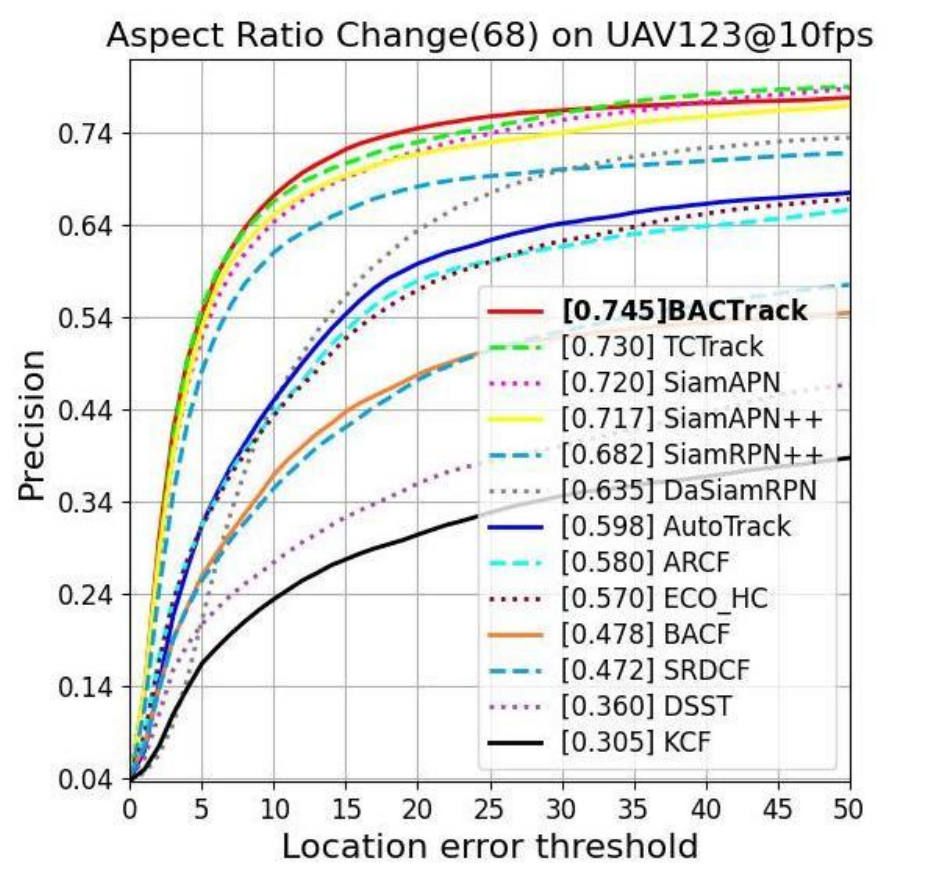}}\hspace{-3mm}
\subfigure{
\includegraphics[width=0.3\linewidth]{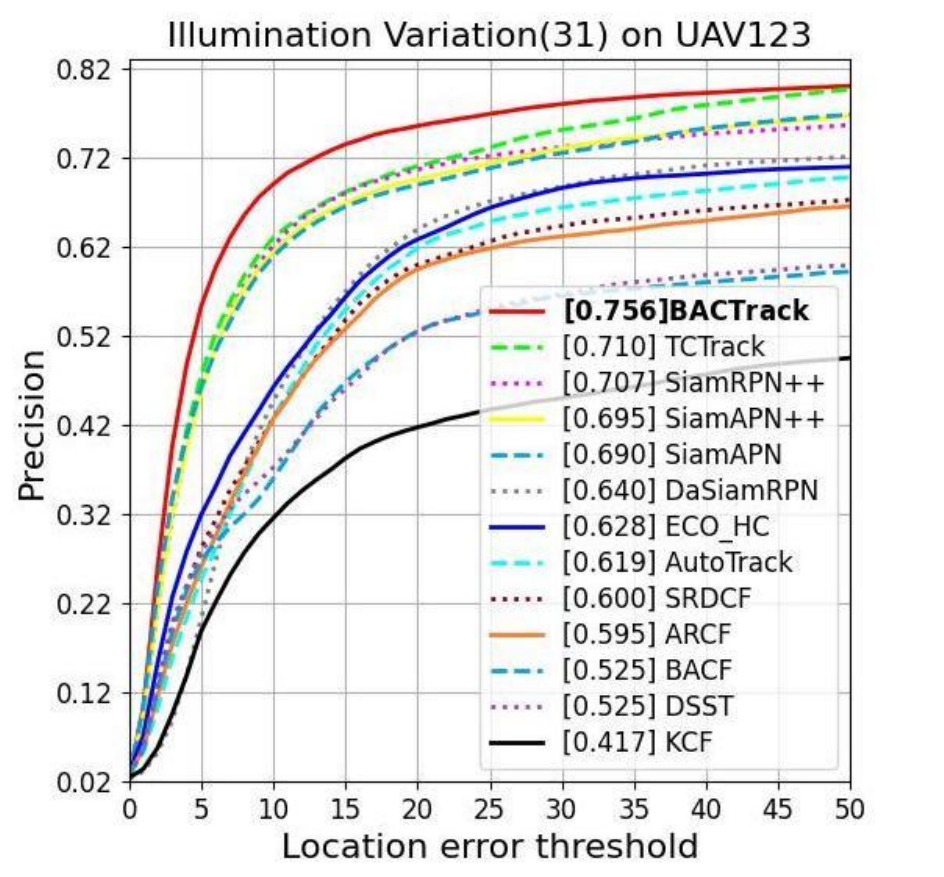}}\hspace{-3mm}
\subfigure{
\includegraphics[width=0.3\linewidth]{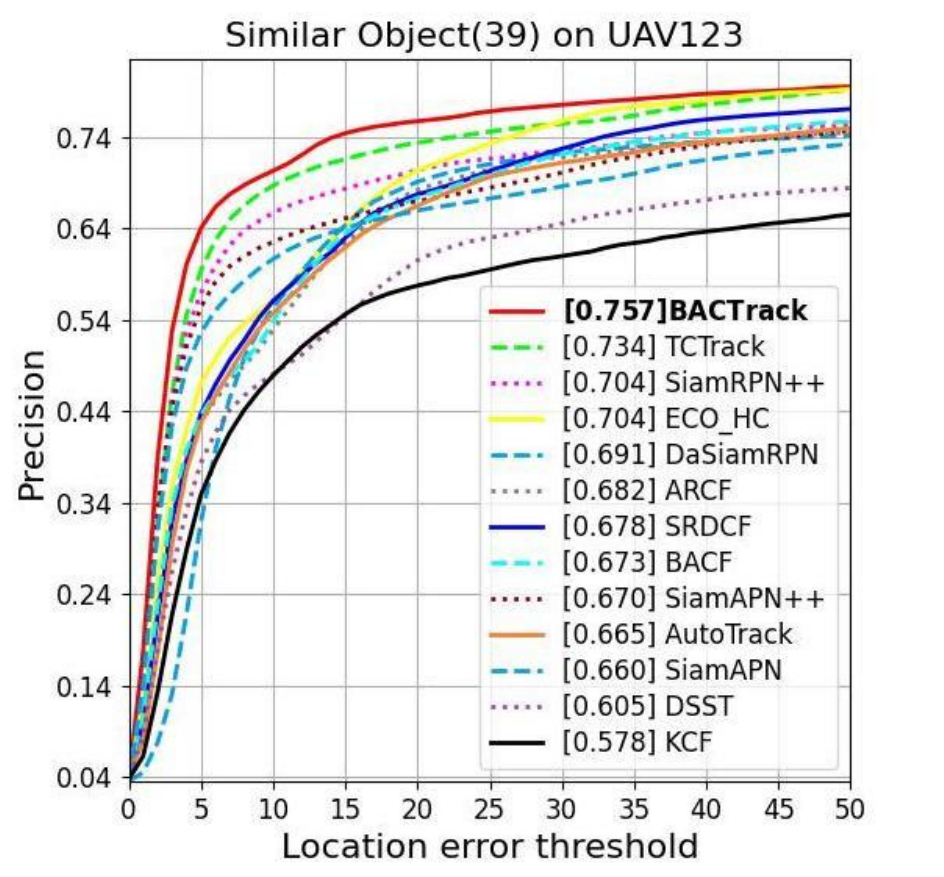}}
\vspace{-1mm}
\caption{\textbf{Attribute-based evaluation on aerial tracking benchmarks.} The experimental results encompass six challenging factors, namely, viewpoint change, deformation, background clutter, aspect ratio change, illumination variation, and similar object.}
\label{fig:Attribute}
\vspace{-2mm}
\end{figure*}

\subsection{Attribute-based Evaluation}
\label{sec:attribute}
To validate the effectiveness of our tracker under complex aerial tracking conditions such as deformation, we also perform attribute-based evaluations.
Fig. \ref{fig:Attribute} provides a precision comparison of six attributes defined across three well-known aerial tracking benchmarks, including viewpoint change, deformation, background clutter, and more.
The comparison with other SOTA trackers, as illustrated in Fig. \ref{fig:Attribute}, proves the robustness of our framework in various challenging conditions.
Notably, our method outperforms state-of-the-art trackers in demanding scenarios such as viewpoint change, illumination variation, and deformation.

Specifically, our method exhibits a $3.1\%$ improvement in handling similar objects compared to the suboptimal TCTrack\cite{TCTrack} and a $1.9\%$ enhancement in addressing viewpoint change compared to the suboptimal SiamAPN++\cite{siamapn++}. These results underscore the effectiveness of our proposed appearance collection.
As our tracker can effectively fuse multiple temporal template features and construct a diverse collection of templates, BACTrack excels in handling similar objects and target deformation scenarios. These challenges are prevalent in real-world UAV object tracking, serving as substantial evidence of the effectiveness of our strategies in the context of UAV tracking.

\subsection{Ablation Study}
\label{sec:abla}
To validate the importance of the components in our tracker, we conduct comprehensive ablation studies, analyzing the influence of different modules, initializations, and parameter settings. 
To ensure the reliability of the comparison, all variants adhere to the same process, including training and parameter settings, with the only variation being the studied module.

\noindent\textbf{Ablations on network components.}
\begin{table}
\begin{center} 
\caption{\textbf{Ablations on network components.} \textbf{MTA}: mixed-temporal attention, \textbf{MHA}: traditional multi-head attention, \textbf{AD}: appearance discriminator, \textbf{PE}: position encoding.}
\renewcommand\arraystretch{1.2}
\setlength{\tabcolsep}{3mm}{
\begin{tabular}{c|cccc|cc}
\toprule
Variant & MHA & MTA & AD & PE & Suc. & Pre.\\
\midrule
Base &\checkmark&           &            &            &0.567  & 0.769\\
Base-AD &\checkmark&           &\checkmark&            & 0.570& 0.775  \\
MTA &          &\checkmark &            &            &0.594  & 0.802 \\
AD &          &\checkmark & \checkmark &            & 0.624 & 0.837 \\\midrule
\textbf{BACTrack} &          &\checkmark & \checkmark & \checkmark &\textbf{0.627} &\textbf{0.843}\\
\bottomrule
\specialrule{0em}{1pt}{1pt}
\end{tabular}}
\label{tab:ablation}
\end{center} 
\vspace{-3mm}
\end{table}
Initially, we validate the effectiveness of the components in our network through ablation studies involving five variants of our framework.
The experimental results of these five variants on DTB70 are summarized in Table \ref{tab:ablation}.
We adopt two attention mechanisms for feature fusion: MHA and MTA, where MHA indicates traditional multi-head attention. 
In variants utilizing MHA, the templates undergo multiple MHAs to independently compute cross-attention of the temporal templates, followed by dimension transformation using $\rm 1\times 1$ convolutions.
AD represents the appearance discriminator, and in the absence of AD, the template is updated every frame. PE denotes the position encoding in Mixed-Temporal Transformer.

We designed a base model 'Base' that utilizes the attention structure of MHA and removes AD, achieving a success rate of $56.7\%$ and a precision of $76.9\%$ (as shown in the first line).
The variant 'MTA' replaces the MHA with the proposed MTA, raising base model performance by about $4.7\%$ in success rate and $4.2\%$ in precision (third line). The significant improvement proves the feasibility of fusing multiple temporal template features through MTA.
In the variant 'AD', we introduce an appearance discriminator, leading to an enhancement of $5.1\%$ in AUC score and $4.3\%$ in precision (fourth line). This outcome underscores the superiority of our template-update strategy, where the AD significantly enhances the quality of introduced temporal information.
The variant 'Base-AD' adds AD to the base model, which further validates this conclusion (second line).
Finally, our complete framework, represented as the variant 'BACTrack', adds position encoding to the MTT, achieving a slight improvement in performance (fifth line).
These experiments show that each part of BACTrack contributes to the enhancement in performance.
\begin{table}
\centering
\caption{\textbf{Ablations on different template-update strategies.} \textbf{AU}: always update, \textbf{STM}: the template update method of STMTrack\cite{Stmtrack}, \textbf{RP}: only reliability principle, \textbf{DP}: only diversity principle, \textbf{AC}: appearance collection.}
\renewcommand\arraystretch{1.2}
\setlength{\tabcolsep}{4mm}{
\begin{tabular}{c|cccc|c} 
\toprule
 & AU & STM & RP & DP & \textbf{AC}\\
\midrule
Suc. &0.594& 0.613 & 0.608 & 0.602 &\textbf{0.627}\\
Pre. &0.802& 0.820 & 0.820 & 0.809 &\textbf{0.843} \\ \bottomrule
\specialrule{0em}{1pt}{1pt}
\end{tabular}}

\label{tab:exp2}
\end{table}
\begin{figure*}
\centering
\includegraphics[width=0.95\linewidth]{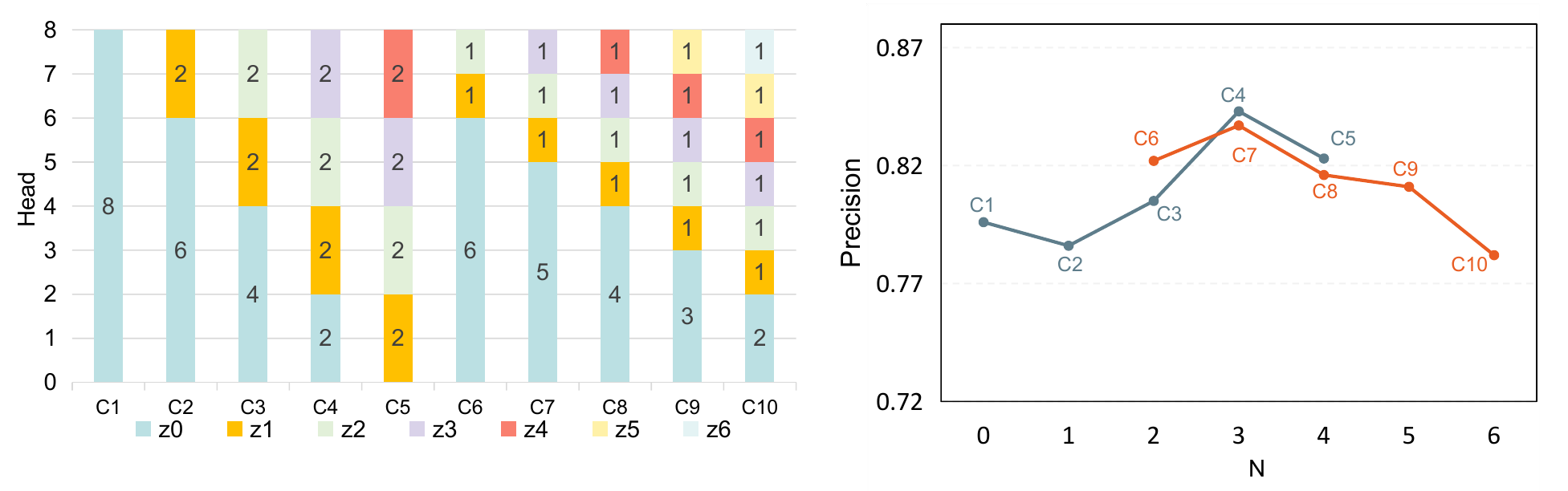}
\vspace{-3mm}
\caption{\textbf{Effect of grouping method and the number of temporal templates $\rm \textbf{N}$ on the performance of BACTrack.} $z_0$ denotes the initial template and $z_1$-$z_6$ denote the temporal templates. The number of heads of the MTA is kept at 8 and other parameters are set the same.} 
\label{fig:group_n}
\end{figure*}

\begin{figure}[t]
\setlength{\belowcaptionskip}{-5cm} 
	\centering  
	\vspace{-0.35cm} 
	\subfigtopskip=2pt 
	\subfigcapskip=-5pt 
	\subfigure[]{
		\label{fig:tau0}
		\includegraphics[width=0.48\linewidth]{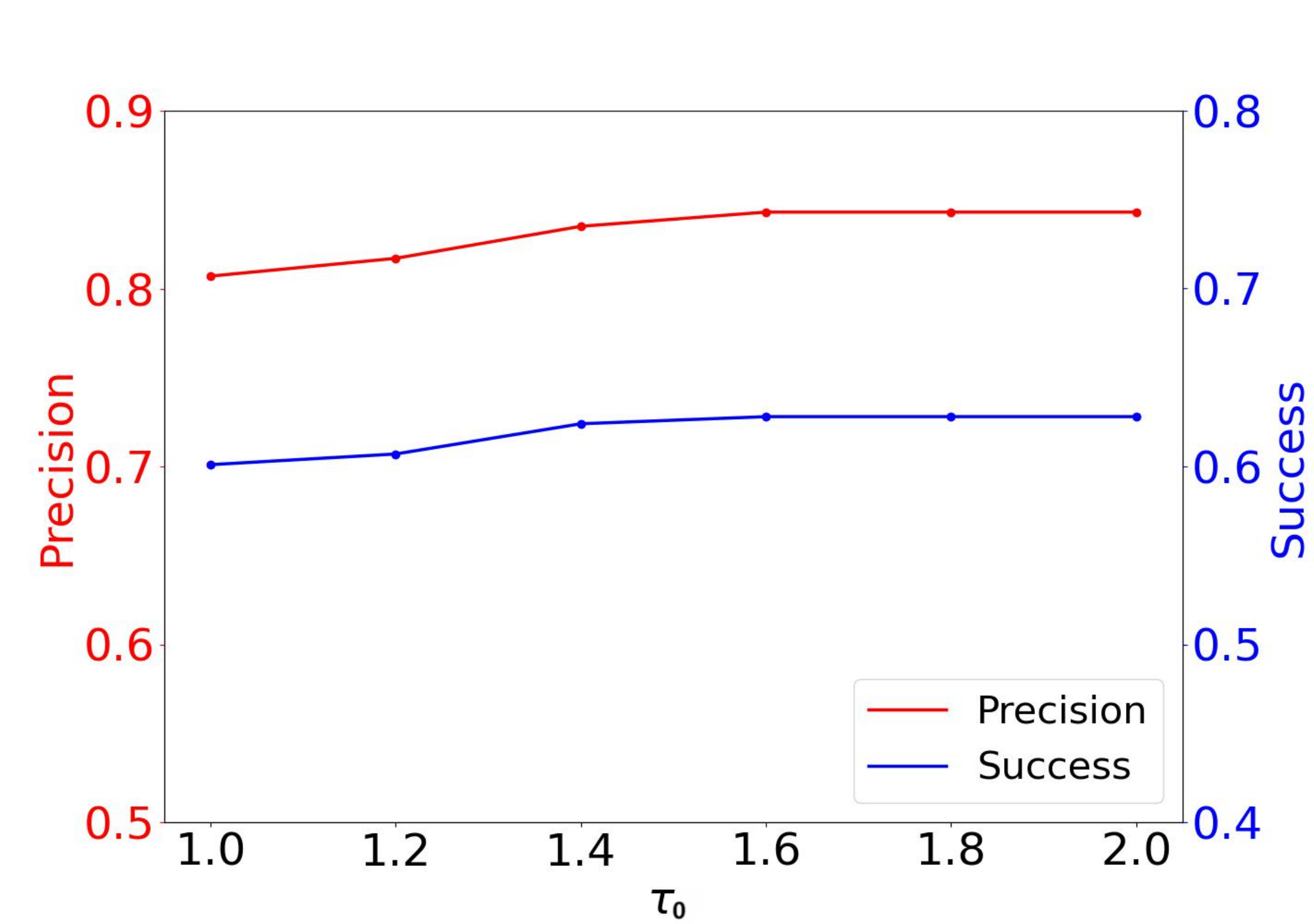}}
	\subfigure[]{
		\label{fig:tau}
		\includegraphics[width=0.48\linewidth]{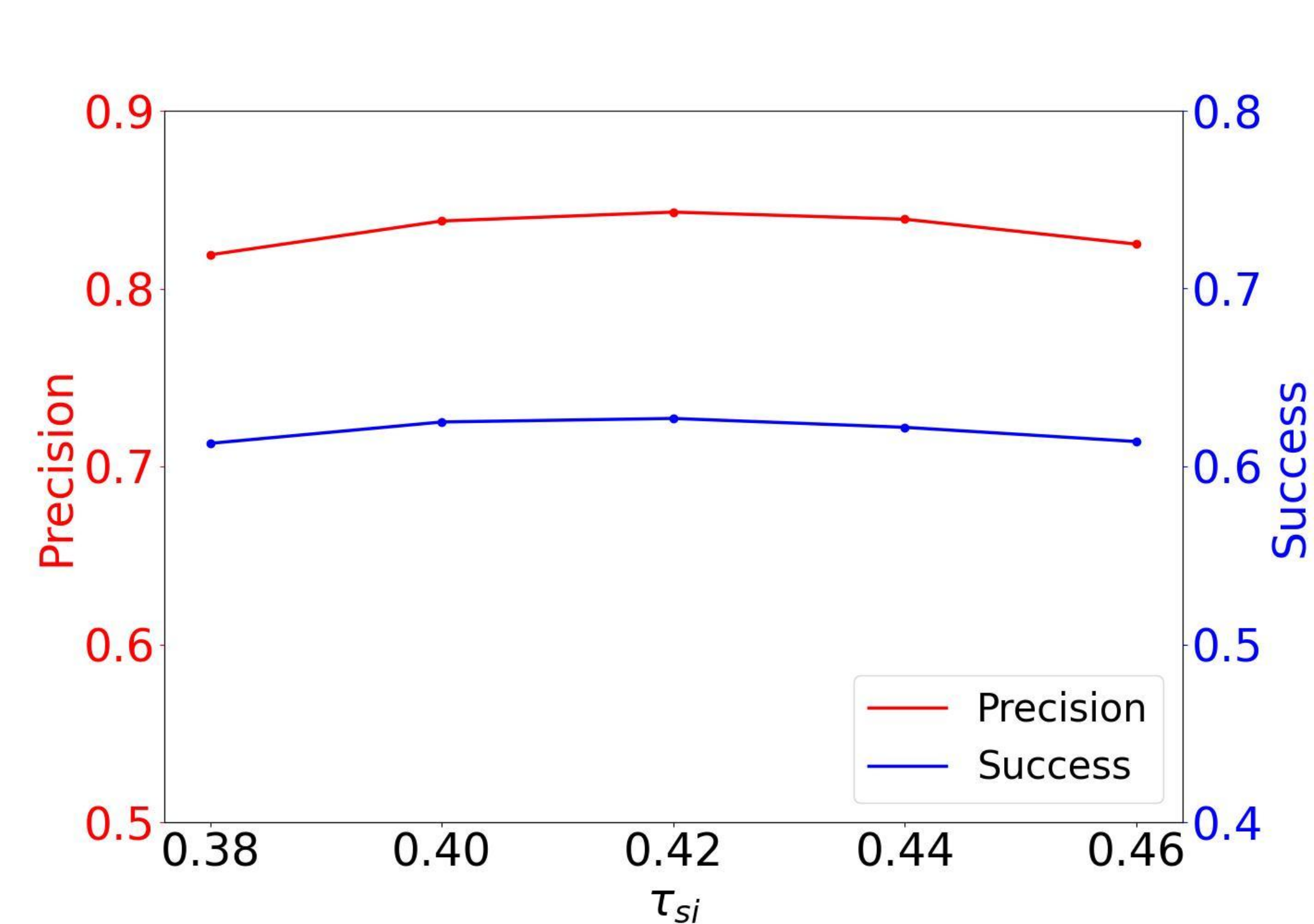}}\\
       \subfigure[]{
		\label{fig:w1}
		\includegraphics[width=0.479\linewidth]{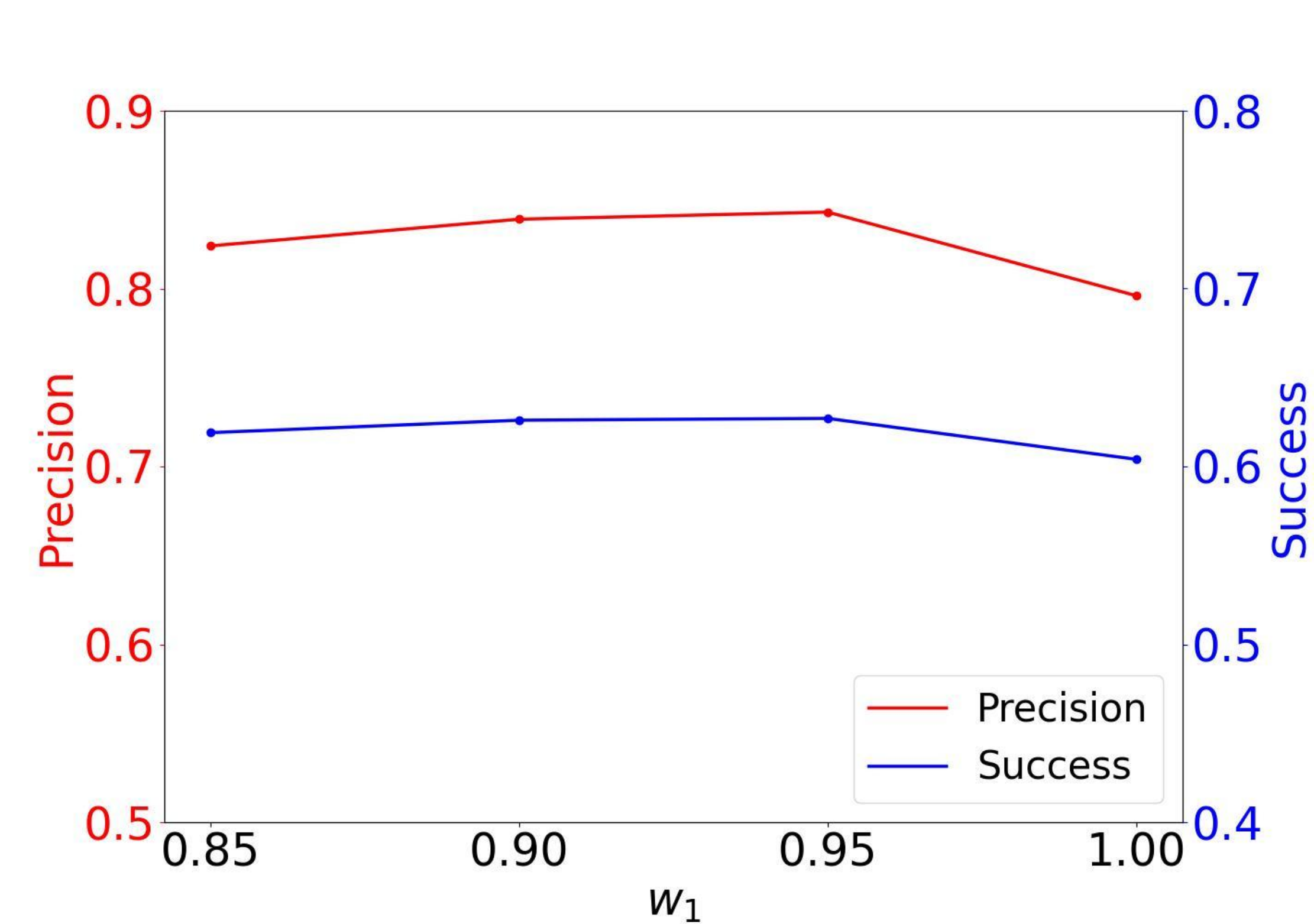}}
	\subfigure[]{
		\label{fig:w2}
		\includegraphics[width=0.479\linewidth]{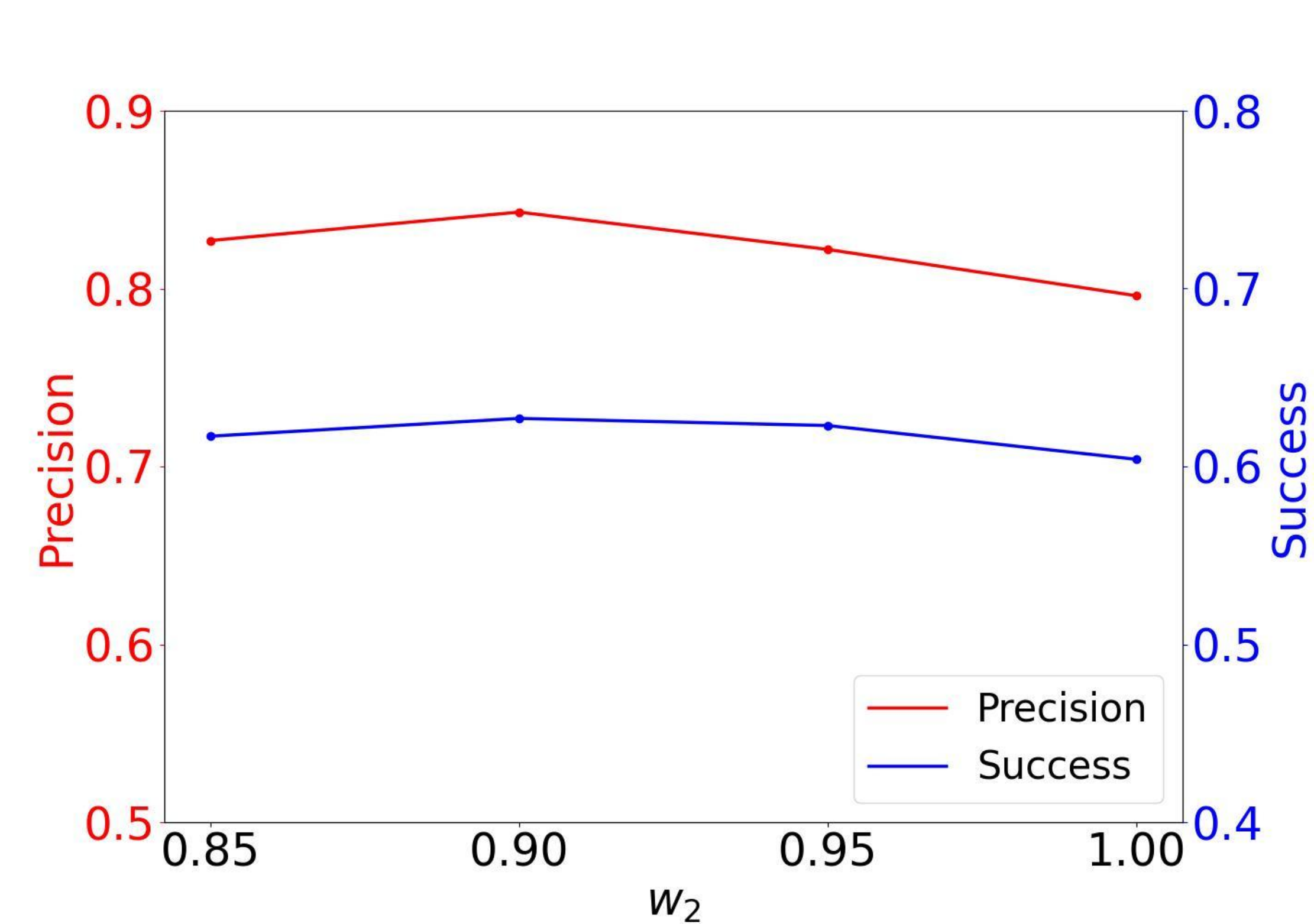}}
        \vspace{-3mm}
	\caption{\textbf{Ablations on parameter settings}: (a) initial confidence threshold $\tau_0 $, (b) similarity threshold $\tau_{\rm si}$, (c) parameter $\rm w_1$, and (d) parameter $\rm w_2$.}
	\label{fig:exp_para}
\end{figure}

\noindent\textbf{Appearance Discriminator.}
To validate the advantages of our template-update strategy in AD, we test the performance of models employing various template-update strategies, as summarized in Table \ref{tab:exp2}. 
The 'AU' (Always Update) strategy involves updating the template in every frame.
'STM' indicates the template update method utilized in STMTrack\cite{Stmtrack}, which divides the historical frames evenly and selects the intermediate frames as the temporal templates.
In addition, our template-update strategy, referred to as 'AC', encompasses two facets: the reliability principle and diversity principle, which we respectively verify through experiments. Specifically, 'RP' denotes the exclusive consideration of the reliability principle, while 'DP' stands for the diversity principle.
As shown in the table, our AC approach achieves the optimal performance, providing compelling evidence for the superiority of our template update mechanism.

\noindent\textbf{Mixed-Temporal Attention.}
We conduct a series of experiments on DTB70 to verify the effect of MTA's structure and scale of templates on the tracking performance.
Initially, we set the number of MTA heads to 8 and modify the number of temporal templates ($\rm N$) from 0 to 6.
To add temporal templates, we replace non-temporal templates with the initial template and design different head grouping methods.
The specific experimental setup and results are shown in Fig. \ref{fig:group_n}, where $z_0$ denotes the initial template and $z_1$-$z_6$ denote the temporal templates. The optimal performance is achieved with configuration C4, which utilizes three temporal templates. These templates are equally divided among the 8 heads, along with the initial frame template.

Then, we maintain the count of three temporal templates while varying the number of MTA heads to 4, 8, and 16.
For each head group, corresponding to a specific template, we utilize 1, 2, or 4 heads for cross-attention calculations.
Remarkably, setting the number of heads to 8 results in a higher AUC ($0.627$) compared to using 4 heads ($0.624$) or 16 heads ($0.586$).
Therefore, BACTrack achieves optimal performance by incorporating three temporal templates, with each template computing attention across 2 heads in parallel.
Decreasing the number of heads necessitates finer-grained features for the templates. And reducing the number of temporal templates results in a lack of diversity within the appearance collection. 
It's important to note that increasing the number of heads and templates may lead to an increase in the model's size and computational requirements. 
Furthermore, increasing the number of temporal templates may lead to feature disorder.

\noindent\textbf{Inference Parameters.}
During the inference phase, we set fixed parameters for the principles of reliability and diversity. 
We conduct experiments to determine these parameters: $\tau_{0}$, $\tau_{\rm si}$, $\rm w_1$, and $\rm w_2$.
The initial confidence threshold $\tau_0 $ determines the template update for the first few frames. Given that the target's motion in these frames is relatively stable and exhibits minimal variation, a large threshold (1.8) is set to maintain the template without updating (as shown in Fig. \ref{fig:tau0}).
Fig. \ref{fig:tau} illustrates the experimental results for $\tau_{\rm si}$, with the highest performance achieved at $\tau_{\rm si} = 0.42$.
When $\tau_{\rm si}$ becomes large, the templates in the collection lack diversity.
Conversely, when $\tau_{\rm si}$ becomes small, the templates in the collection differ significantly from the current target appearance.
Appropriate $\tau_{\rm si}$ effectively filters out templates that are too similar to those in the appearance collection, allowing for better perception of target changes.

In addition, the impact of the parameters $\rm w_1$ and $\rm w_2$ on the performance of BACTrack is presented in Fig. \ref{fig:w1} and Fig. \ref{fig:w2}. 
We set $\rm w_1$ and $\rm w_2$ as $0.95$ and $0.9$, respectively.
When $\rm w_1$ and $\rm w_2$ become large, the threshold $\tau_{t}$ becomes large, resulting in a reduced update rate of the templates. This limitation prevents the full utilization of temporal information. Conversely, when $\rm w_1$ and $\rm w_2$ become small, the appearance discriminator fails to filter out insufficient temporal information, leading to performance degradation.

\vspace{-3 mm} 
\subsection{Extension on General Object Tracking}
\begin{figure}
\centering
\includegraphics[width=1.0\linewidth]{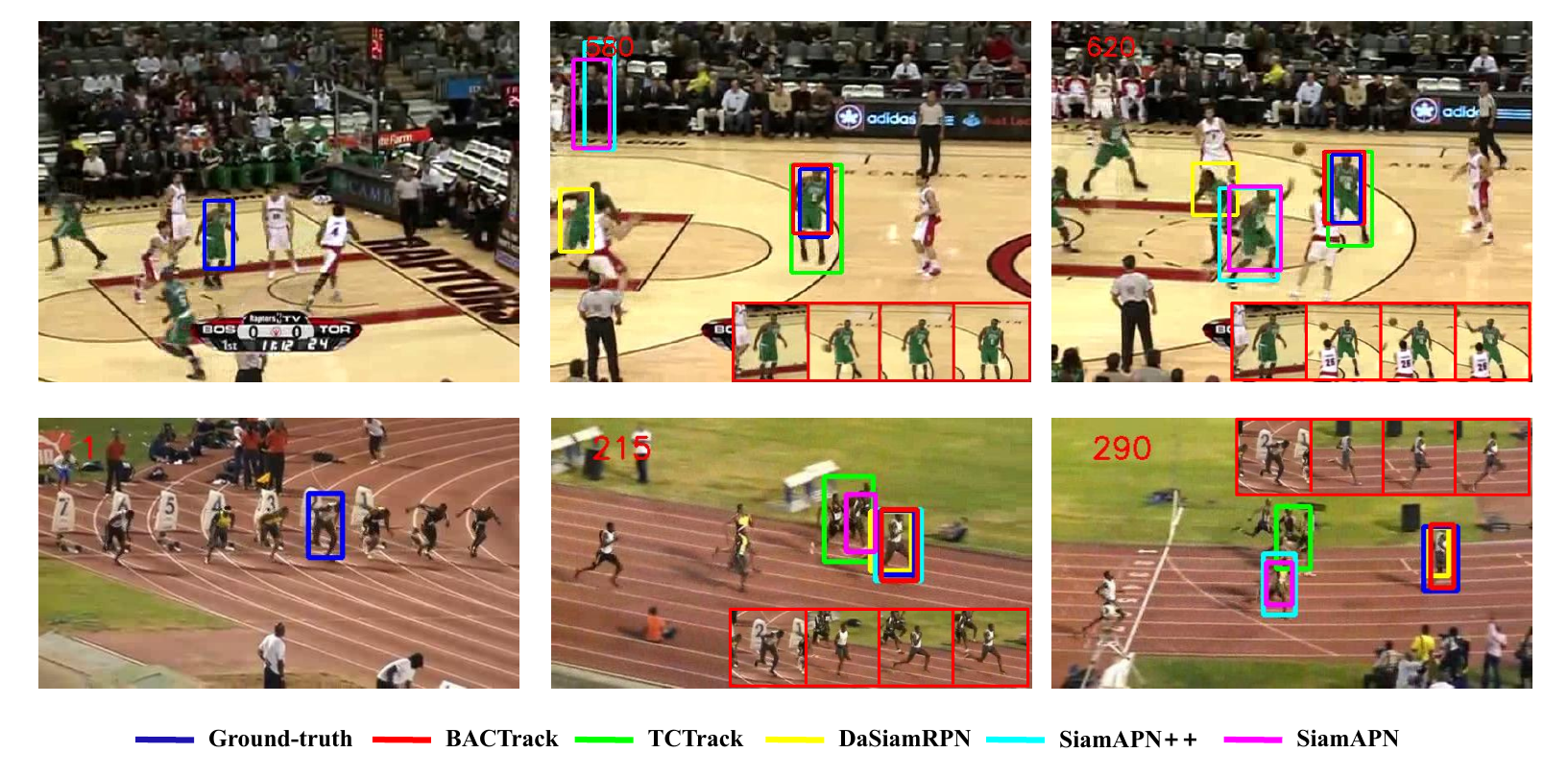}
\setlength{\abovecaptionskip}{-0.3cm} 
\vspace{-3mm}
\caption{\textbf{Qualitative results on OTB2015.} The appearance
collection is marked with red boxes.}
\label{fig:otb_vis}
\end{figure}

To further validate the efficacy of the appearance collection in coping with background interference and drastic changes in target appearance, we conduct additional experiments on OTB2015.
OTB2015 is a renowned benchmark for object tracking, comprising 100 short-term tracking sequences that encompass 11 common challenges, including background clutter.
In Table \ref{tab:otb}, we report the performance of BACTrack and four other SOTA lightweight trackers on OTB2015. Additionally, we provide attribute-based evaluation results for background clutter (BC) and illumination change (IV).
Fig. \ref{fig:otb_vis} intuitively presents the qualitative comparison of BACTrack with other top-performing trackers on OTB2015.
Our BACTrack surpasses state-of-the-art lightweight trackers.
Since our appearance collection can efficiently utilize temporal information and store changing target templates, BACTrack achieves outstanding performance in solving background clutter and other challenges.
\begin{table}
\centering
\caption{\textbf{Overall performance of BACTrack and other SOTA trackers on OTB2015.} \textbf{BC}: background clutter, \textbf{IV}: illumination change. The best performance is marked in \textbf{bold}.}
\renewcommand\arraystretch{1.2}
\begin{tabular}{c|cc|cc|cc}
\toprule
\multirow{2}{*}{Trackers} & \multicolumn{2}{c|}{Overall} & \multicolumn{2}{c|}{BC} & \multicolumn{2}{c}{IV} \\ \cline{2-7}
                          & Suc.         & Pre.         & Suc.       & Pre.      & Suc.       & Pre.      \\ \midrule
DaSiamRPN\cite{DaSiamRPN}                 & 0.552        & 0.765        & 0.514      & 0.589     & 0.556      & 0.799     \\
SiamAPN\cite{SiamAPN}                   & 0.636        & \textbf{0.838}        & 0.560       & 0.752     & 0.646      & 0.840      \\
SiamAPN++\cite{siamapn++}                 & 0.634        & 0.824        & 0.554      & 0.723     & 0.629      & 0.798     \\
TCTrack\cite{TCTrack}                   & 0.615        & 0.825        & 0.563      & 0.753     & 0.603      & 0.786     \\
\midrule
\textbf{BACTrack (Ours)}                  & \textbf{0.639}        &\textbf{0.838}        & \textbf{0.589}      & \textbf{0.790}    & \textbf{0.656}      & \textbf{0.856}     \\ \bottomrule
\end{tabular}
\label{tab:otb}
\end{table}
\section{Failed Cases and Limitations}
\begin{figure}
\centering
\includegraphics[width=1.0\linewidth]{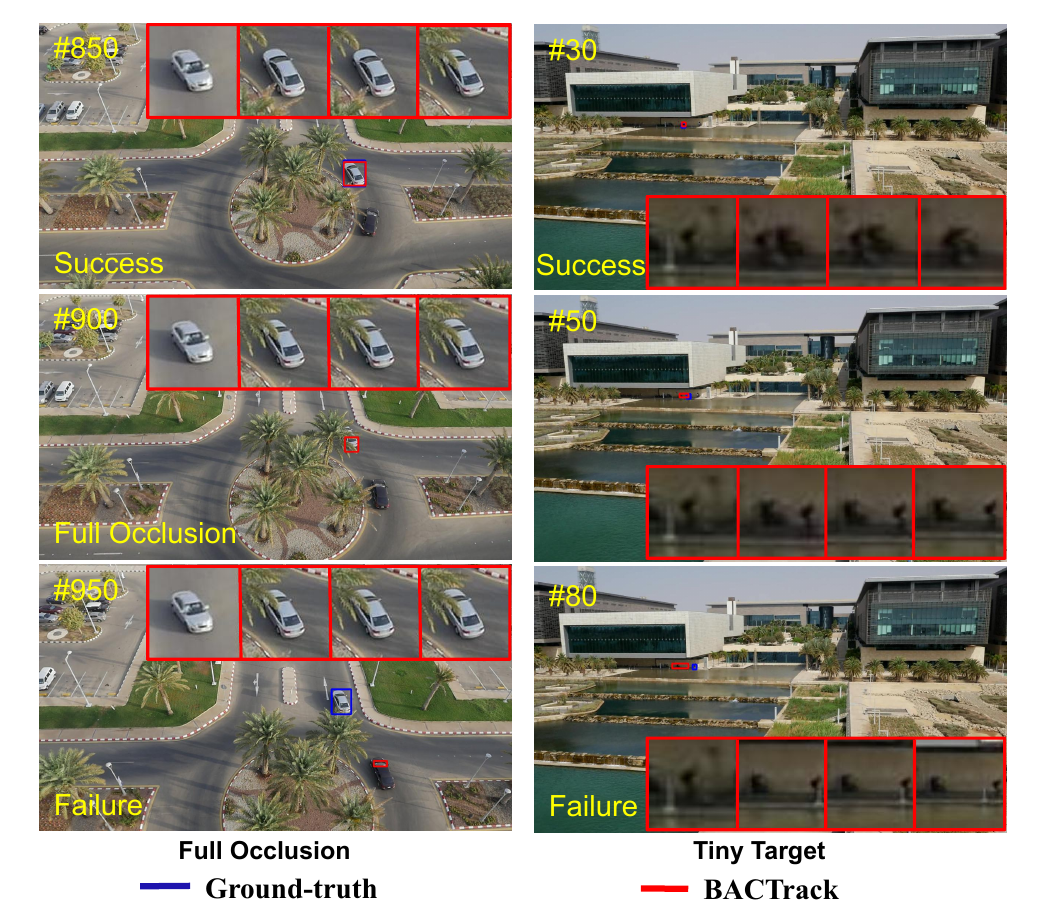}
\setlength{\abovecaptionskip}{-0.3cm} 
\vspace{-3mm}
\caption{\textbf{Failed Cases.} BACTrack may fail to localize the target when the target is extremely small or appears to be completely obscured.}
\label{fig:fail}
\end{figure}

Despite BACTrack's excellent performance in aerial tracking, there are instances of failure that provide opportunities for further improvement, as shown in Fig. \ref{fig:fail}.
In aerial views, targets can be very small, often leading to an insufficient collection of appearances to effectively distinguish the target from its background.
Additionally, scenarios involving occlusion by other objects or complex interactions frequently challenge the tracker, sometimes resulting in errors. In particular, even with the right templates in the collection, our tracker may struggle to reacquire the target after occlusion.
Due to lightweight considerations, our tracker is not equipped with a global detector to cope with target out of view and extended periods of full occlusion. Addressing these challenges, particularly improving the tracker's ability to handle occlusions in UAV scenarios, is a key direction for our subsequent research.

\section{Conclusion}
In this work, we present BACTrack, a lightweight tracking framework aimed at establishing an adaptive and dynamic collection of target templates during online tracking. BACTrack efficiently integrates multi-template features, enhancing the tracking robustness significantly. 
A key component of BACTrack is the Mixed-Temporal Transformer (MTT) module, devised for feature fusion. This module builds relationships between the search region and multiple target templates through a mixed-temporal attention mechanism.
Additionally, the framework incorporates an appearance discriminator, complemented by an adaptive online template-update strategy for efficient template maintenance.
Building upon these innovations, we build a target appearance collection as templates, enabling the preservation of temporal information and enhancing adaptability to complex target appearance variations.
Comprehensive experiments have shown that BACTrack performs well on four authoritative aerial tracking benchmarks while running at 87.2 FPS on a GPU. Additional evaluations on edge devices further validate the deployability of BACTrack on real drone platforms.

{\small
\bibliographystyle{IEEEtran}
\bibliography{reference}
}

\end{document}